\documentclass[sigconf]{acmart}

\usepackage{booktabs}

\usepackage{makecell}
\usepackage[ruled,vlined]{algorithm2e}
\usepackage{amsthm}
\usepackage{amsmath}
\usepackage{amsfonts}
\usepackage{multirow}
\usepackage{enumitem}
\usepackage{array}
\usepackage{subfigure}
\usepackage{graphicx}
\usepackage{url}
\usepackage{xspace}
\usepackage{upgreek}
\usepackage{color}
\usepackage{microtype}

\newcommand{\hide}[1]{} 
\newcommand{\modelname}{MuKGE\xspace}
\newcommand{\modelnamernn}{MuKGE (RNN)\xspace}
\newcommand{\modelnamersn}{MuKGE (RSN)\xspace}
\newcommand{\modelnameformer}{MuKGE (TF)\xspace}

\newcommand{\fbtwo}{FB15K-237\xspace}

\newcommand{\wnrr}{WN18RR\xspace}
\newcommand{\yg}{YAGO3-10\xspace}
\newcommand{\dbp}{DBP15K\xspace}
\newcommand{\dbplp}{DBP15K-LP\xspace}
\newcommand{\wdfivem}{Wikidata5M\xspace}
\newcommand{\dbpfivem}{DBpedia5M\xspace}
\newcommand{\wikidbp}{WikiDBP10M\xspace}

\newcommand{\wqsp}{WebQuestionsSP\xspace}

\newcommand{\myparagraph}[1]{\smallskip\noindent\textbf{#1.}\xspace}

\newtheorem{definition}{Definition}

\SetCommentSty{mycommfont}

\AtBeginDocument{%
  \providecommand\BibTeX{{%
    \normalfont B\kern-0.5em{\scshape i\kern-0.25em b}\kern-0.8em\TeX}}}

\copyrightyear{2023} 
\acmYear{2023} 
\setcopyright{acmlicensed}\acmConference[KDD '23]{Proceedings of the 29th ACM SIGKDD Conference on Knowledge Discovery and Data Mining}{August 6--10, 2023}{Long Beach, CA, USA}
\acmBooktitle{Proceedings of the 29th ACM SIGKDD Conference on Knowledge Discovery and Data Mining (KDD '23), August 6--10, 2023, Long Beach, CA, USA}
\acmPrice{15.00}
\acmDOI{10.1145/3580305.3599397}
\acmISBN{979-8-4007-0103-0/23/08}

\begin{document}


\title[Joint Pre-training and Local Re-training on Multi-source Knowledge Graphs]{Joint Pre-training and Local Re-training: Transferable Representation Learning on Multi-source Knowledge Graphs}


\author{Zequn Sun}
\author{Jiacheng Huang}
\affiliation{
  \department{State Key Laboratory for Novel Software Technology}
  \institution{Nanjing University \country{China}}
}
\email{zqsun.nju@gmail.com}
\email{jchuang.nju@gmail.com}

\author{Jinghao Lin}
\author{Xiaozhou Xu}
\author{Qijin Chen}
\affiliation{
    \institution{Alibaba Group \country{China}}
}
\email{jinghaolin.ljh@alibaba-inc.com}
\email{heixia.xxz@alibaba-inc.com}
\email{qijin.cqj@alibaba-inc.com}

\author{Wei Hu}
\authornote{Wei Hu is the corresponding author.}
\affiliation{
    \department{State Key Laboratory for Novel Software Technology}
    \department{National Institute of Healthcare\\ Data Science}
    \institution{Nanjing University \country{China}}
}
\email{whu@nju.edu.cn}

\renewcommand{\shortauthors}{Zequn Sun et al.}

\begin{abstract} 
In this paper, we present the ``joint pre-training and local re-training'' framework for learning and applying multi-source knowledge graph (KG) embeddings.
We are motivated by the fact that different KGs contain complementary information to improve KG embeddings and downstream tasks.
We pre-train a large teacher KG embedding model over linked multi-source KGs and distill knowledge to train a student model for a task-specific KG. 
To enable knowledge transfer across different KGs, we use entity alignment to build a linked subgraph for connecting the pre-trained KGs and the target KG.
The linked subgraph is re-trained for three-level knowledge distillation from the teacher to the student, i.e., feature knowledge distillation, network knowledge distillation, and prediction knowledge distillation, to generate more expressive embeddings. 
The teacher model can be reused for different target KGs and tasks without having to train from scratch.
We conduct extensive experiments to demonstrate the effectiveness and efficiency of our framework.
\end{abstract}



\maketitle

\section{Introduction}\label{sec:introduction}
\begin{figure}[!t]
\centering
\includegraphics[width=0.8\linewidth]{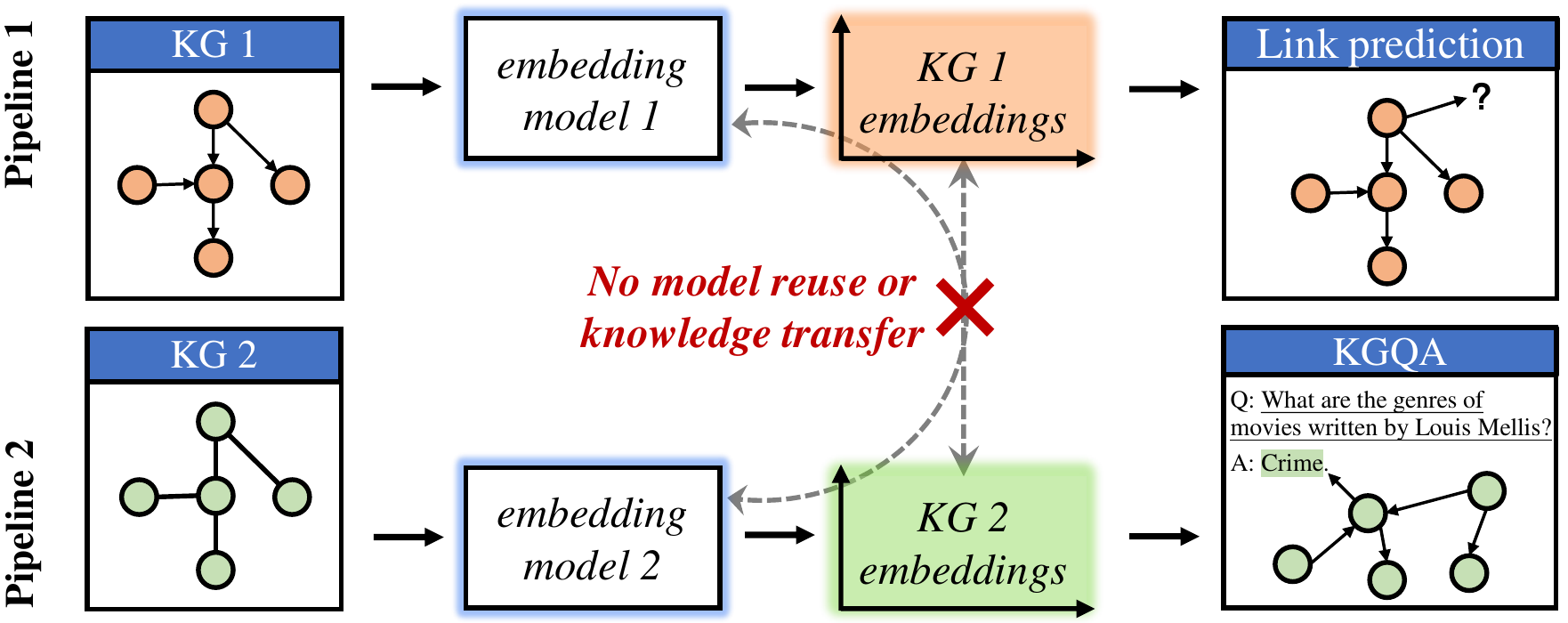}
\caption{Conventional KG embedding and its application.}
\label{fig:vs}
\end{figure}

To structure human knowledge and make it machine-processable for downstream tasks,
people have invested significant efforts in developing large-scale knowledge graphs (KGs),
such as Freebase~\cite{Freebase}, Wikidata \cite{Wikidata}, DBpedia \cite{DBpedia}, YAGO \cite{YAGO} and NELL \cite{NELL}, by harvesting structured facts from the web data or human-annotated corpora.
In recent years, a typical paradigm for applying KGs to downstream tasks usually consists of two modules~\cite{TKDE_KGE,KG_survey}, as shown in Figure~\ref{fig:vs}.
The first module is to learn the embeddings of the target KG from scratch,
and the second is to inject these embeddings into the task. 
The two modules can be processed step-by-step \cite{MultiHopQA_KGE}, or simultaneously optimized to take advantage of their deep interactions \cite{ERNIE}.
A practical challenge here lies in how to learn high-quality KG embeddings that yield improved performance on a downstream task.
The target task always expects KG embeddings to be more expressive, capture more semantics, and cover more entities.

However, in realistic scenarios, KGs are far from complete by nature \cite{KGCompleteness}.
Embeddings learned from sparse and incomplete KGs would lack expressiveness.
As a result, downstream tasks frequently suffer from a knowledge deficit.
To cope with this issue, researchers have explored two lines of solutions. 
One is to complete a KG by inferring missing triplets (a.k.a link prediction) based on the learned embeddings \cite{TransE,RGCN,ConvE,TuckER,ATTH_acl20}. 
However, embedding learning in this solution still suffers from the inherent incompleteness of the KG and is unable to utilize auxiliary knowledge from other sources.
The other is to merge different KGs by linking identical entities (a.k.a. entity alignment) in embedding space \cite{MTransE,JAPE,GCN_Align,MuGNN,AliNet}. 
However, existing entity alignment studies are mainly concerned with new entity matches rather than the shared knowledge in KGs.

Recent progress has demonstrated that joint representation learning over multiple KGs can leverage the shared knowledge to generate expressive embeddings \cite{MTransE,LinkNBed,RSN,EnsembleKGC,KBDistillation}. 
However, we still face challenges in the following aspects. 
First, introducing other background KGs into joint learning to help the target KG embedding would {cost more time and resources} than training the target KG alone.
Second, joint representation learning is unable to reuse the pre-trained models. 
When confronted with other target KGs and tasks,
we have to train the multiple KGs from scratch.
Third, joint learning on multiple KGs suffers from {insufficient knowledge transfer}. 
Due to the different scales of KGs,
when the joint model converges, the embeddings of the target KG may be underfitting.
Thus, while joint learning can leverage shared knowledge, it is incapable of fully exploiting the potential of multi-source KGs.
This method is not efficient in real-world applications, 
which calls for a new paradigm for transferable representation learning on multi-source KGs and applying it to downstream tasks.

Then, an interesting question is raised: Is the ``pre-training and fine-tuning'' paradigm applicable to multi-source KGs?
It first pre-trains a large model using a massive text or image corpus, 
and then fine-tunes the model for a target task \cite{BERT,BiT}.
We argue that, the data properties of KGs are quite different from those of texts and images, which makes pre-training and fine-tuning inapplicable.
The basic unit of text is word.
Two tokens with the same literal representation refer to the same word (although a word may have different meanings in different contexts).
The basic unit of images is pixel.
Two pixels with the same gray value in different images refer to the same color.
We do not need to consider word alignment or pixel alignment in pre-training and fine-tuning.
However, an object in KGs may have different symbolic representations and structures.
We have to consider how to embed multi-source KGs in a joint space and how to transfer knowledge to the task.

In this paper, we propose a \textit{pre-training} and \textit{re-training} framework for improving the target KG embedding via knowledge transfer from linked multi-source KGs. 
It jointly pre-trains a large representation learning (teacher) model over multiple KGs and then trains a small (student) model via knowledge distillation for downstream tasks on the target KG, e.g., link prediction in one KG \cite{TransE}, and multi-hop KG question answering \cite{MultiHopQA_KGE}. 
The research question in our pre-training is how to capture the shared knowledge background KGs, whereas that in re-training is how to transfer useful knowledge to the student model on the target KG.
Entity alignment is used to bridge the target and background KGs, and enables knowledge transfer.
The contributions of this paper are threefold:

\begin{itemize}
\item Our key contribution is the \textit{pre-training and re-training} framework for KG embedding learning and application. It first pre-trains an embedding model over multi-source background KGs. Then, given a task-specific KG, it employs the pre-trained model as a teacher and distills useful knowledge to train a student embedding model. It can offer improved quality-speed trade-offs to enhance KG embeddings.

\item We define and explore the design space of path encoders. Specifically, we explore three solutions based on recurrent neural networks (RNN) \cite{LSTM}, recurrent skipping networks (RSN) \cite{RSN}, and Transformer \cite{Transformer}. For effective and efficient knowledge transfer between the heterogeneous structures in background and target KGs, we propose re-training with the linked subgraph by multi-level knowledge distillation.

\item For evaluation, in addition to conventional link prediction (LP), we also consider two knowledge transfer settings, i.e., the joint LP setting and our pre-training and re-training setting.
In each setting, we consider two groups of background KGs that have heterogeneous or similar schemata to the target KG. 
We also conduct an experiment on multi-hop KG question answering.
The experiments show the effectiveness, efficiency and generalization of our framework.
\end{itemize}

\section{Problem Statement}\label{sec:statement}
We start with the definition of KGs.

\begin{definition}[Knowledge graph]
A KG $\mathcal{K}$ is defined as a three-tuple $\mathcal{K}=(\mathcal{E},\mathcal{R},\mathcal{T})$, 
where $\mathcal{E}$ and $\mathcal{R}$ are the sets of entities and relations, respectively. 
$\mathcal{T} \subseteq \mathcal{E}\times \mathcal{R}\times\mathcal{E}$ is the set of relational triplets.
\end{definition}

There are many publicly available KGs.
Different KGs may share a set of identical entities. 
Driven by the Linked Open Data (LOD) project, there are a large number of multi-source KGs that are linked by identical entities (i.e., entity alignment).
For the sake of generality, we consider entity alignment between two KGs and it can be extended to more KGs without difficulty by leveraging the transitivity of equivalence.
Formally, the definition is given below:

\begin{definition}[Entity alignment]
Given two different KGs $\mathcal{K}_1=(\mathcal{E}_1,\mathcal{R}_1,\mathcal{T}_1)$ and $\mathcal{K}_2=(\mathcal{E}_2,\mathcal{R}_2,\mathcal{T}_2)$,
their entity alignment $\mathcal{A}_{1,2}$ is defined as a set of identical entities,
i.e., $\mathcal{A}_{1,2}=\{(e_1,e_2)\,|\,e_1\equiv e_2, e_1\in \mathcal{E}_1, e_2\in \mathcal{E}_2\}$,
where $\equiv$ denotes semantic equivalence.
\end{definition}

Motivated by LOD, we consider linked multi-source KGs as background for knowledge transfer. 
The definition is as follows:

\begin{definition}[Linked multi-source KGs] 
Linked multi-source KGs refer to a collection of KGs, i.e., $\mathcal{M}=\{\mathcal{K}_1^{(b)}, \dots, \mathcal{K}_n^{(b)}\}$, 
where each KG is linked with at least one other KG by entity alignment.
That is, for $\forall \, \mathcal{K}_i^{(b)} \in \mathcal{M}$, $\exists \, \mathcal{K}_j^{(b)} \in \mathcal{M} \ \text{and}\ i \neq j$ such that $|\mathcal{A}_{i,j}|>0$.
\end{definition}

In our setting, we require each KG to have entity alignment with at least one other KG, such that the KG is able to participate in knowledge transfer.
However, identical entities in different KGs usually have heterogeneous relational structures \cite{AliNet}.
For example, DBpedia (DBP) does not have the ``\textit{grandfather}'' relation.
It uses (X, \textit{father}, Y) and (Y, \textit{father}, Z) to express the fact that Z is the \textit{grandfather} of X.
But, Wikidata (WD) has the ``\textit{grandfather}'' relation, and the fact is represented as a triplet (X, \textit{grandfather}, Z).
The successful knowledge transfer should capture the relational pattern:
\begin{equation}\label{path}
	\text{X} \xrightarrow[]{\textit{WD:grandfather}} \text{Z}
	\Leftarrow 
	\text{X} \xrightarrow[]{\textit{DBP:father}} \text{Y} \xrightarrow[]{\textit{DBP:father}} \text{Z},
\end{equation} 
which motivates us to consider long-term relational paths in linked KGs for embedding learning.
Relational triplets can be regarded as a special case of paths, 
but they may not capture long-term relational semantics \cite{RSN}.
Other solutions, such as triplet-based \cite{TransE} and GNN-based \cite{RGCN} models, are not a good choice for our setting.

\begin{definition}[Relational path]
A relational path is defined as an entity-relation chain that starts with a head entity and ends with a tail entity, where relations appear alternately. 
For brevity, we use $(x_1, x_2, \cdots, x_t)$ to denote a path with length $t$, where $t\geq 3$ is an odd number.
$x_i$ is an entity if $i$ is an odd number, otherwise it is a relation.
\end{definition}

Existing transferable KG embedding models only focus on capturing entity similarity \cite{OpenEA} while ignoring relational reasoning across different KGs.
Our problem setting is defined as follows:

\begin{definition}[Knowledge transfer from linked multi-source KGs to the target KG]
Given linked multi-source KGs $\mathcal{M}$, target KG $\mathcal{K}^{(t)}$, and their entity alignment $\mathcal{A}$,
it pre-trains a teacher embedding model (if one does not exist), denoted by $M_{\text{teach}}$, over $\mathcal{M}$.
It then distills $M_{\text{teach}}$ to learn a model, denoted by $M_{\text{stu}}$, for $\mathcal{K}^{(t)}$.
\end{definition}

Our framework does not require that: (i) the target KG is linked to each background KG, and (ii) each KG is linked to all others.
It only requires each KG to have entity alignment with at least one of the background KGs to jump-start the knowledge transfer.
The teacher model can be reused for different target KGs.
\begin{figure}[t]
\centering
\includegraphics[width=0.9\linewidth]{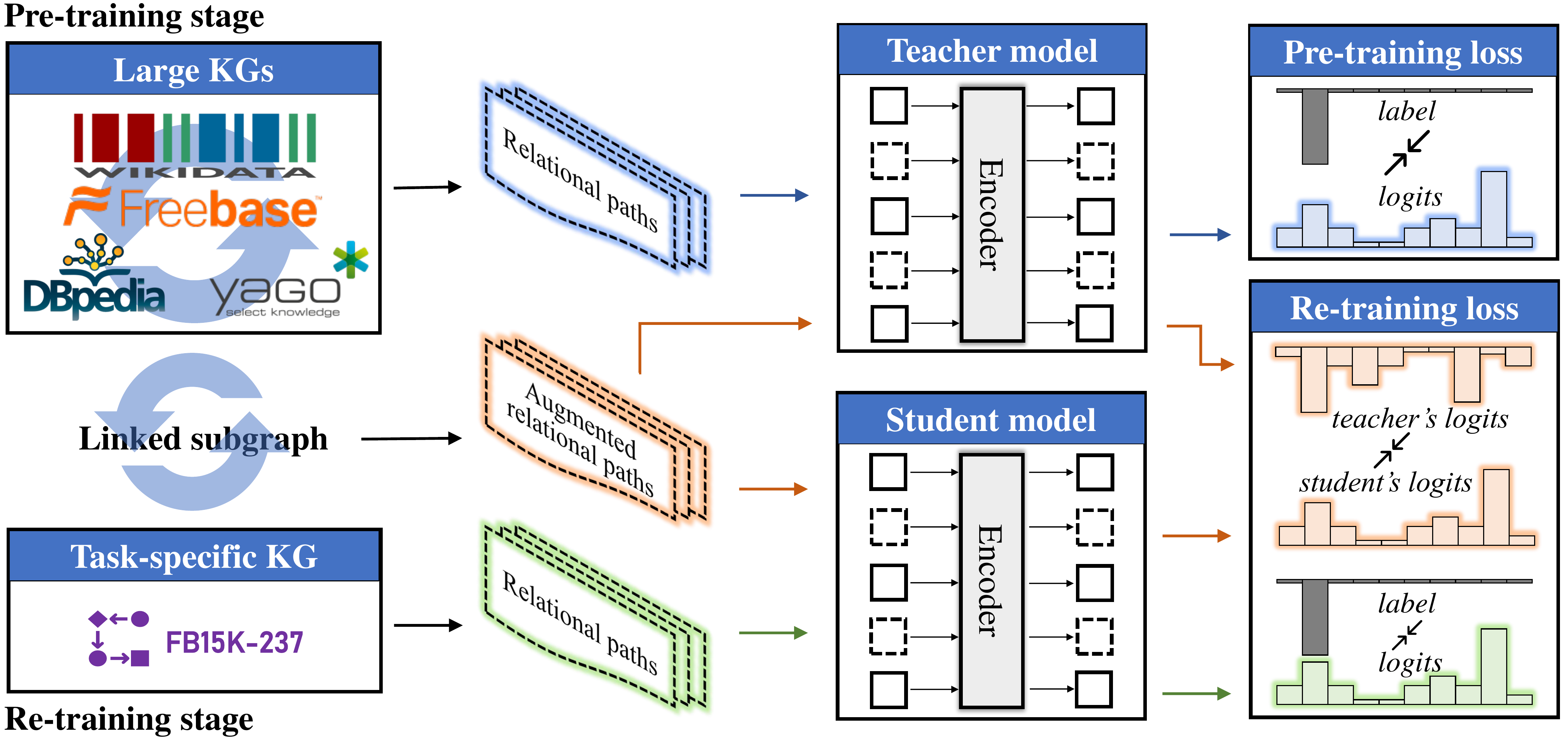}
\caption{Illustration of pre-training and re-training for transferring knowledge from multi-source KGs to the target KG.}
\label{fig:pipeline}
\end{figure}

\section{Pre-training and Re-training} \label{sect:method}

\subsection{Framework Overview}
The framework is depicted in Figure~\ref{fig:pipeline}, which consists of two phases:

\myparagraph{Joint pre-training} 
We assume that at least one background KG is provided for pre-training. 
Background KGs are usually larger than the target KG and contain entity alignment with it.
We pre-train a teacher embedding model on background KGs with self-supervised tasks like entity prediction and relation prediction.

\myparagraph{Local re-training}
Given the teacher model and entity alignment, 
we train a student model for the target KG by training over the linked subgraph and distilling useful knowledge from the teacher.

\subsection{Path Encoder}
Given a relational path $(x_1, x_2, \cdots, x_t)$, we first retrieve the input representations of the entities and relations, and then feed the path into an encoder with $L$ layers stacked. 
The input embeddings are denoted by $(\mathbf{x}_1^0, \mathbf{x}_2^0, \cdots, \mathbf{x}_t^0)$, 
and the output of the $L$-th layer is
\begin{equation}
    (\mathbf{x}_1^L, \mathbf{x}_2^L, \cdots, \mathbf{x}_t^L) = \texttt{Encoder}\,(\mathbf{x}_1^0, \mathbf{x}_2^0, \cdots, \mathbf{x}_t^0).
\end{equation}

\myparagraph{Design space}
We hereby discuss the key principled designs of path encoders.
For path modeling, any sequence encoder can be used. 
We consider three typical networks as the backbone of the path encoder, RNN \cite{LSTM}, RSN \cite{RSN} and Transformer \cite{Transformer}.

\smallskip\noindent\textbf{RNN}
is a popular family of neural networks for modeling sequence data. 
It can also model the entity-relation paths in KGs.
Given an input relational path, $(\mathbf{x}_1^0, \mathbf{x}_2^0, \cdots, \mathbf{x}_t^0)$,
a RNN sequentially reads each element and outputs a corresponding hidden state. 
At time step $t$, the output hidden state, denoted by $\mathbf{h}_t$, is calculated as
\begin{equation}
\label{eq:rnn}
\mathbf{h}_t = f(\mathbf{h}_{t-1}, \mathbf{x}_t),
\end{equation}
where $f(\cdot,\cdot)$ is a multilayer perceptron used for combination. $\mathbf{h}_{0}=\mathbf{0}$.
Due to the problem of vanishing gradients in vanilla RNNs, 
we use the Long Short-Term Memory (LSTM) network \cite{LSTM}.
It introduces an internal state $\mathbf{c}_t$ to help capture the information for long periods of time.
Its output hidden state is calculated as
\begin{align}\label{eq:lstm}
\mathbf{c}_t = \mathbf{f}_t \odot (\mathbf{c}_{t-1}) + \mathbf{i}_t \odot f(\mathbf{h}_{t-1}, \mathbf{x}_t), \, \, \, \mathbf{h}_t = \mathbf{o}_t \odot \tanh(\mathbf{c}_t),
\end{align}
where $\mathbf{f}_t$, $\mathbf{i}_t$ and $\mathbf{o}_t \in [0,1]^d$ denote three gates to control the information passing,
and $\odot$ denotes the element-wise product.

\smallskip\noindent\textbf{RSN} \cite{RSN} is a RNN variant for KG embedding. 
It improves LSTM by adding a residual connection to link the subject entity embedding and its subsequent relation embedding along a path,
such that the model is able to directly capture relational semantics and deliver them to the object entity.
Its output is calculated as
\begin{align}
\label{eq:rsn}
\mathbf{h}'_t = 
\begin{cases}
\mathbf{h}_t & x_t\in \mathcal{E} \\
\mathbf{W}_1 \mathbf{h}_t + \mathbf{W}_2 \mathbf{x}_{t-1} & x_t\in\mathcal{R}
\end{cases},
\end{align}
where $\mathbf{h}'_t$ denotes the output hidden state of RSN at time step $t$, and $\mathbf{h}_t$ is the corresponding hidden state of LSTM in Eq.~(\ref{eq:lstm}). 
$\mathbf{W}_1$ and $\mathbf{W}_2$ are two weight matrices for combination.

\smallskip\noindent\textbf{Transformer} \cite{Transformer} is widely acknowledged as a powerful network for modeling sequential data.
Its key component is the multi-head self-attention that can capture the dependence between any two elements in the input sequence:
\begin{equation}\label{eq:transformer}
\mathbf{h}_t = \texttt{self-attention}\,(\mathbf{x}_1^0, \mathbf{x}_2^0, \cdots, \mathbf{x}_t^0).
\end{equation}

\subsection{Joint Pre-training over Background KGs}
In the pre-training phase, 
we first sample relational paths for each background KG by random walks, and then augment these paths with entity alignment to generate cross-KG paths as training data.
The paths are fed into a sequence encoder to get the output representations for multi-task self-supervised learning.

\myparagraph{Path sampling with random walk}
Random walk on KGs is a widely used method to generate paths.
Unlike existing work that only uses entity sequences \cite{RDF2Vec_swj},
we also consider relations in KGs.
Enumerating all the possible walks leads to a large number of paths.
The massive scale of the data would result in a huge time cost.
Hence, we only repeat several random walks (e.g., $n=2$) from a start point.
Before that, we add reverse triplets to the KG to ensure that each entity has an outgoing edge.
Given a triplet $(s,r,o)$, we denote the reverse version of $r$ by $r[\text{reverse}]$, and get the reverse triplet $(o,r[\text{reverse}],s)$.
To guarantee that each triplet would be visited in the random walk and speed up path sampling, we let each walk start with the object $o$ of a triplet ($s,r,o$), such that we have a path of length 3 before starting walking.
Next, we choose from the neighbors of entity $o$ based on their weights.
The sampled neighbor $o'$ and the connecting relation $r'$ are appended to the triplet, resulting in a path of length $5$, i.e., ($s,r,o,r',o'$).
Then, we can continue to walk from the current start point $o'$.
Thus, to get a path of length $l$ ($l\ge 3$), the number of steps for each walk is $\frac{l-3}{2}$.

\myparagraph{Path augmentation}\label{sect:path_aug}
When given multiple background KGs,
we use path augmentation to transfer triplet knowledge between KGs.
The idea is that identical entities from different KGs should hold the relations of each other.
We use two strategies for path augmentation.
The first is entity replacement.
Given a relational path in a KG, we replace one of its entities with the counterpart entity in another background KG. 
Let us consider two background KGs $\mathcal{K}^{(b)}_1$ and $\mathcal{K}^{(b)}_2$ along with their entity alignment $\mathcal{A}_{1,2}$.
Given a relational path $(s,r,o,r_1,o_1)$ in $\mathcal{K}^{(b)}_1$, and an aligned entity pair $(o, o')\in\mathcal{A}_{1,2}$, we can generate a new relational path $(s,r,o',r_1,o_1)$.
The augmented path in this way connects entities from different KGs.
However, it does not connect the relations of two KGs.
The second strategy, i.e., path concatenation, deals with this issue.
It generates a long path by concatenating two different paths where the object entity of one path is aligned with the subject of another path.
For example, given a relational path $(s,r,o)$ in $\mathcal{K}^{(b)}_1$ and $(s',r',o')$ in $\mathcal{K}^{(b)}_2$, and an aligned entity pair $(o, s')\in\mathcal{A}_{1,2}$, we can generate two new paths $(s,r,o,r',o')$ and $(s,r,s',r',o')$.

\myparagraph{Multi-task self-supervised objectives}
We consider two self-supervised tasks: (masked) relation prediction and (masked) entity prediction.
The key idea is that each entity or relation in a relational path can be predicted based on its contexts.
Computing the Softmax or BCE loss for entity or relation prediction is time-consuming and even not acceptable in real scenarios due to the large scale of candidate space. 
Hence, we adopt the noise-contrastive estimation (NCE)~\cite{NCE} to compute the prediction loss.
NCE only requires a small number of negative samples to approximate the probability distribution.
Given an input path $(s,r,s',r',o)$, take entity prediction as an example.
The probability of predicting entity $o$ given the current context $c_o$ is defined as
\begin{equation}\label{eq:prob}
p(o\,|\,c_o) = \sigma(\mathbf{o}\cdot \mathbf{c}_{o}),
\end{equation}
where $\sigma (\cdot)$ denotes sigmoid, and $\mathbf{o}$ is the input representation of $o$.
$\mathbf{c}_{o}$ denotes the output representation of the context in the encoder.
In RNN and RSN, the context of an entity or relation is its predecessor elements in the path.
In Transformer, the context refers to all other elements.
Specifically, the NCE loss of entity prediction is
\begin{equation}\label{eq:loss}
\mathcal{L}_{\text{ent}}(o) = - \log p(o\,|\,c_o)
+ \sum_{i=1}^{k} \mathbb{E}_{\tilde{e}_i\sim Q(\tilde{e})} \big[1- \log{p(\tilde{e}_i\,|\,c_o)} \big],
\end{equation}
where $k$ denotes the number of negative samples. 
The negative entity $\tilde{e}_i$ is randomly sampled from the rectified probability distribution: $Q(\tilde{e}) \propto q(\tilde{e})^{\frac{3}{4}}$, 
where $q(\tilde{e})$ denotes the frequency of entity $\tilde{e}$ in the KG.
The loss for relation prediction, denoted by $\mathcal{L}_{\text{rel}}(r)$, is defined in a similar way.
Put it all together and we get the following loss for the input relational path $(s,r,s',r',o)$:
\begin{equation}\label{eq:overall_loss}
\mathcal{L}_{\text{KG}} = \mathcal{L}_{\text{ent}}(s) + \mathcal{L}_{\text{rel}}(r) + \mathcal{L}_{\text{ent}}(s') + \mathcal{L}_{\text{rel}}(r') +\mathcal{L}_{\text{ent}}(o).
\end{equation}

Depending on the prediction task (entity or relation prediction), the negative samples can be either negative entities or relations. 

\subsection{Local Re-training for Knowledge Transfer}
We propose three strategies to distill knowledge from the input representations, intermediate layers, and output logits from the teacher model to improve the student model. 
But the heterogeneity and independence of different KGs present difficulties for knowledge transfer.
Fortunately, we can obtain some entity alignment of multiple KGs thanks to the LOD project \cite{LOD}.
We use such seed entity alignment to build a bridge between multi-source KGs.
Inspired by recent work \cite{JOS}
we define the linked subgraph, the common ``space'' shared by the background KGs and the target KG.
The linked subgraph plays two important roles in our knowledge transfer. 
First, it is used for re-training to let the embeddings of the target KG move towards the background KGs' representation space to get benefits.
Second, it builds a unified label space for aligning the prediction logits of local entities from the teacher and student models.

\begin{definition}[Linked subgraph] 
Given the background KG $\mathcal{K}^{(b)}$ and the target KG $\mathcal{K}^{(t)}$, let $\mathcal{E}^{(b)}$ be the set of background entities in their entity alignment $\mathcal{A}$, the linked subgraph is denoted by a set of triplets $\mathcal{G}=\{(s,r,o)\in \mathcal{T}^{(b)}\,|\, s \in \mathcal{E}^{(b)} \ \text{or}\ o \in \mathcal{E}^{(b)}\}$.
\end{definition}

We have $|\mathcal{E}^{(b)}| \le |\mathcal{E}^{(t)}|$. 
In most cases, only a small portion of entities have alignment with the background KGs. 
However, the subgraph could still be very large, and even larger than the target KG, i.e., we may have $|\mathcal{G}| > |\mathcal{T}^{(t)}|$.
Using the whole linked subgraph in re-training leads to a lot of time overhead.
To reduce re-training time while ensuring the knowledge transfer performance, we propose a sampling method.
Given a budget $b$, the goal is to sample a small linked subgraph $\mathcal{G}' \subseteq \mathcal{G}$ such that $|\mathcal{G}'| \le b$.
The key idea is motivated by the finding \cite{OpenEA} that more entity alignment pairs contribute more to aligning the embedding spaces of different KGs, and the entities with denser relations also contribute more than long-tail entities.
Therefore, we give preference to the triplets, denoted by $\mathcal{G}''=\{(s,r,o)\in \mathcal{T}^{(b)} \,|\, s \in \mathcal{E}^{(b)} \ \text{and}\ o \in \mathcal{E}^{(b)}\}$, whose subject and object entities both have links.
If $|\mathcal{G}''|>b$, we can randomly sample $b$ triplets from $|\mathcal{G}''|$ to build the linked subgraph.
If $|\mathcal{G}''|<b$, we need to sample additional $b'=|\mathcal{G}''|-b$ triplets.
We consider the triplets whose entities are popular in the linked subgraph $\mathcal{G}$.
We first count the frequency of each entity appearing in $\mathcal{G}''$. 
The popularity of a triplet is defined as the sum of the frequencies of its subject and object entities. 
We do not select the top-$b'$ ``popular'' triplets because including too many these triplets would result in the hubness issue that hurts embedding learning \cite{OpenEA}.
Instead, we use weighted random sampling to get $b'$ triplets from $\mathcal{G}-\mathcal{G}''$ based on triplet popularities.
Finally, we get the linked subgraph $\mathcal{G}'$, which participates in the re-training process.

The linked subgraph has been pre-trained in the teacher model.
It also participates in re-training for knowledge transfer.
We generate and augment the relational paths of both the linked subgraph and target KG,
and feed them into a student model.
Re-training seeks to preserve relational semantics in the target KGs while benefiting from the useful knowledge from background KGs and the teacher model.
We have three-level knowledge distillation (KD) methods.

\myparagraph{Feature knowledge distillation}
We consider the input representations of entities as feature knowledge. 
We expect that each entity in the linked subgraph should have similar representations in the student and teacher models,
i.e., the linked entity in the target KG should have similar representations with its counterpart in the linked subgraph. 
A naive solution is to initialize the input representations of the entities in the linked subgraph as their pre-trained input embeddings in the teacher model.
However, considering that the embedding dimensions of the two models may be different, we introduce a learnable matrix for embedding transformation and minimize the embedding distance. 
The loss is given as follows:
\begin{equation}\label{eq:input_kd_loss}\footnotesize
\mathcal{L}_{\text{feat}} = \frac{1}{|\mathcal{E}'|} \sum_{e\in \mathcal{E}'} \text{MSE}(\mathbf{W}_{\text{feat}}\mathbf{e}^0,\mathbf{e}^0_{\text{teach}}) + \frac{1}{|\mathcal{A}'|} \sum_{(e_1, e_2)\in \mathcal{A}'} \text{MSE}(\mathbf{e}^0_1,\mathbf{e}^0_2),
\end{equation}
where $\mathcal{E}'$ denotes the entity set of the final linked subgraph $\mathcal{G}'$, and $\mathcal{A}'$ is the entity alignment between the linked subgraph and target KG.
$\text{MSE}(\cdot,\cdot)$ denotes the mean squared error, and $\mathbf{W}_{\text{feat}}$ is the transformation matrix.
$\mathbf{e}^0$ denotes the input representation of entity $e$ in the student model, 
while $\mathbf{e}^0_{\text{teach}}$ is the corresponding pre-trained input embedding in the teacher model.

\myparagraph{Network knowledge distillation}
The second type of knowledge is the trainable network parameters.
The intermediate-level knowledge from the teacher model layers can guide the learning of student layers \cite{FitNets}. 
Let $\Theta$ be the parameter set. The loss is 
\begin{equation}\label{eq:network_kd_loss}
\mathcal{L}_{\text{net}} = \frac{1}{|\Theta|} \sum_{\theta\in \Theta}\text{MSE}(\mathbf{W}_{\theta}\mathbf{\theta}, \mathbf{\theta}_{\text{teach}}),
\end{equation}
where $\mathbf{\theta}$ is a parameter in the student while $ \mathbf{\theta}_{\text{teach}}$ is the corresponding parameter in the teacher.
$\mathbf{W}_{\theta}$ is the transformation matrix.

\myparagraph{Prediction knowledge distillation}
Let $\mathcal{E}'$ and $\mathcal{P}'$ be the sets of entities and relational paths of the linked subgraph, respectively.
The linked subgraph builds a common label space for entity prediction such that we can align the prediction probabilities of the two models.
Take an input path $(s,r,s',r',o)\in\mathcal{P}'$ as an example,
and we would like to predict $o$ with $\mathcal{E}'$ as the candidate space.
We feed the path into the teacher model and normalize the prediction probability through a softmax function over $\mathcal{E}'$ as follows:
\begin{equation}\label{eq:teach_pr}
Pr_{\text{teach}}(o\,|\,c_o) = \frac{p_{\text{teach}}(o\,|\,c_o)}{\sum_{e\in \mathcal{E}'}p_{\text{teach}}(e\,|\,c_e)},
\end{equation}
where $p_{\text{teach}}(o\,|\,c_o)$ denotes the prediction probability that can be calculated by Eq.~(\ref{eq:prob}). 
As the number of entities in the linked subgraph (i.e., $|\mathcal{E}'|$) is limited, here we can compute the softmax.
We use the input embeddings and output representations in the teacher model for computation.
The probability distribution $Pr_{\text{teach}}$ is used as a guide information for training the student model. 
The pre-trained teacher model is frozen and the gradient does not back to the teacher. 
In a similar way, we can feed the path into the student model to get the logits as follows:
\begin{equation}\label{eq:stu_pr}
Pr_{\text{stu}}(o\,|\,c_o) = \frac{p_{\text{stu}}(o\,|\,c_o)}{\sum_{e\in \mathcal{E}'}p_{\text{stu}}(e\,|\,c_o)},
\end{equation}
where we use the embeddings and output representations of the student model for computing the probability.
We use the KL divergence to measure the probability distribution of entity prediction:
\begin{equation}\label{eq:kd_prob_loss}
\mathcal{L}_{\text{prob}} = \text{KL}\big(Pr_{\text{teach}}(o\,|\,c_o), Pr_{\text{stu}}(o\,|\,c_o)\big).
\end{equation}

\myparagraph{Overall loss} 
The overall loss for knowledge distillation is
$\mathcal{L}_{\text{KD}} = \mathcal{L}_{\text{feat}} + \alpha \cdot \mathcal{L}_{\text{net}} + \beta \cdot \mathcal{L}_{\text{prob}}$,
where $\alpha$ and $\beta$ are hyper-parameters.

\subsection{Put It All Together}

\myparagraph{Pre-training process} 
We merge background KGs into a set of relational paths $\mathcal{P}$ via path sampling and augmentation.
Given a path encoder such as RSN~\cite{RSN} or Transformer~\cite{Transformer}, we train the teacher model using $\mathcal{P}$ based on mini-batch stochastic gradient optimization.
We first randomly initialize the model parameters, including the input representations of KGs.
Then, in each step, we randomly sample a path batch and feed it into the encoder.
With the output, we compute the pre-training loss Eq.~(\ref{eq:overall_loss}) and the corresponding gradient to update model parameters.
The training is terminated when it reaches the maximum number of epochs.

\myparagraph{Re-training process} 
The re-training includes an additional knowledge distillation objective.
To jump-start knowledge distillation, we first sample the linked subgraph based on the entity links between the target KG and background KGs.
The augmented relational paths of the linked subgraph are also used for embedding learning in the student model. 
Instead of joint training, our implementation optimizes the self-supervised prediction loss and knowledge distillation loss alternately.
Re-training is terminated using early-stop based on the validation performance of the target task.

\myparagraph{Application} 
There are two typical ways of applying the student model to a target task.
One is to feed the embeddings from the student model into the task model (e.g., a QA model).
The other is to jointly optimize the student model (with knowledge transfer from the teacher) and the task model in a multi-task learning framework.

\section{Experiments and Results}\label{sect:setup}

\subsection{Setup}
We consider two KG embedding tasks.
\textbf{Link prediction} attempts to fill in the missing object or subject entity given an incomplete triplet in the target KG \cite{TransE}.
\textbf{Multi-hop KGQA} predicts the answer entity from a target KG given a natural language query \cite{MultiHopQA_KGE}. 

\myparagraph{Research questions}
We attempt to answer the following research questions through the experiments and analyses:

\begin{itemize}
\item \textbf{RQ1}. What about the effectiveness (Sect.~\ref{sect:results}, \ref{sect:similar_results}) and efficiency (Sect.~\ref{sect:time}) of pre-training and re-training? 
\item \textbf{RQ2}. How do the three path encoders perform? (Sect.~\ref{sect:exp_encoder}) What about the multi-level knowledge distillation? (Sect.~\ref{sect:ablation})
\item \textbf{RQ3}. Why is joint representation learning not effective enough? How does our framework outperform it? (Sect.~\ref{sect:joint_training})
\item \textbf{RQ4}. What about the generalization performance of our framework in downstream tasks, e.g., KGQA? (Sect.~\ref{sect:exp_qa})
\item \textbf{RQ5}. Is any background KG helpful? What enables successful or causes negative knowledge transfer? (Sect.~\ref{sect:case})
\end{itemize}

\myparagraph{Environment and implementation}
The experiments were carried out on a GPU server with an Intel(R) Xeon(R) Gold 6248 CPU (2.50GHz), 256GB memory, and four NVIDIA Tesla V100 GPUs.
The operating system is Ubuntu 18.04. 
We make the source code, dataset, and running scripts available at our GitHub repository\footnote{\url{https://github.com/nju-websoft/MuKGE}}.

\subsection{Datasets}
We consider two different dataset settings:

\noindent\textbf{Datasets in different schemata.}
In this setting, the background KGs and the target KG are organized by different schemata (i.e., ontologies).
This is a real and challenging setting. 
We choose two large-scale KGs, i.e., Wikidata \cite{Wikidata} and DBpedia \cite{DBpedia},
for pre-training a teacher model to improve link prediction on three widely-adopted datasets \fbtwo \cite{TransE}, \wnrr \cite{ConvE} and \yg \cite{YAGO3-10}.
We report the statistics in Table~\ref{table:datasets} of Appendix~\ref{app:lp_dataset}.
WordNet is a commonsense KG while Freebase and YAGO are open-domain.
\yg is much larger than \fbtwo and \wnrr.
As for the background KG Wikidata, we use its subset \wdfivem \cite{wd5m}.
For DBpedia, we use its English version.
We further merge \wdfivem and \dbpfivem as a larger KG \wikidbp by 
merging aligned entities as a ``virtual'' node.
Their statistics are given in Table \ref{tab:backgroundKG}. 

\myparagraph{Datasets in similar schemata}
To further evaluate knowledge transfer, we build a new dataset \dbplp based on the entity alignment benchmark \dbp~\cite{JAPE}.
\dbp has three entity alignment settings, i.e., ZH-EN (Chinese-English), JA-EN (Japanese-English) and FR-EN (French-English). 
Each setting contains $15,000$ entity alignment pairs. 
We merge English triplets in the three settings and get four linked KGs (i.e., ZH, JA, FR and EN) as shown in Figure~\ref{fig:dbp15k_dataset_graph} of Appendix~\ref{app:dbp_lp}. 
For link prediction in each KG, we randomly split $80\%$, $10\%$ and $10\%$ of triplets as training, validation and test data, respectively. 
\dbp is extracted from infobox-based triplets of DBpedia. Its relations are not mapped to a unified ontology. 
Hence, the four KGs in \dbplp are organized by similar schemata rather than a unified ontology.

\subsection{Link Prediction Settings}
To thoroughly investigate knowledge transfer between different KGs, we consider three evaluation settings for link prediction:

\begin{itemize}

\item \textbf{Conventional link prediction (LP)}.
This is the conventional setting that does not consider knowledge transfer.

\item \textbf{Joint LP (JointLP)}.
It is a transfer learning setting. 
We merge background KGs and the target KG's training triplets as a unified graph (by merging aligned entities) to train a model and evaluate it on the test triplets of the target KG.

\item \textbf{LP via pre-training and re-training~{(PR4LP)}}.
In this setting,
we distill knowledge from a teacher model trained~with background KGs to train a student link prediction model.

\end{itemize}

\begin{table}[!t]
\centering
\caption{\label{tab:backgroundKG}Statistics of background KGs for pre-training.}
\resizebox{0.999\linewidth}{!}{
\begin{tabular}{lrcrc}
\hline
KG & \# Entities & \# Relations & \# Triplets & \# Alignment \\
\hline
\wdfivem~\cite{wd5m} & $4,594,485$ & \ \ \ $822$ & $21,354,359$ & \multirow{2}{*}{$1,462,188$} \\
\dbpfivem (ours) & $5,382,945$ & \ \ \ $653$ & $18,093,812$ & \\
\hline
\wikidbp (ours) & $8,739,436$ & $1,481$ & $39,448,171$ & $-$ \\
\hline
\end{tabular}}
\end{table}

\subsection{Baselines and Evaluation Metrics} 

Our variants are denoted by \modelnamernn, \modelnamersn and \modelnameformer, respectively.
To our best knowledge, there is no existing work for JointLP and PR4LP. 
We select four representative KG embedding models as baselines. 
They are TransE \cite{TransE}, RotatE \cite{RotatE}, ConvE \cite{ConvE}, and TuckER \cite{TuckER}.
We do not compare with recent GCN-based models like CompGCN \cite{CompGCN} because their implementations are not scalable to large background KGs with 256GB memory.

\begin{table}[!t]
\centering
\caption{\label{tab:links}\# Entity alignment with background KGs.}
\resizebox{0.8\linewidth}{!}{{\small
\begin{tabular}{lrrr}
\hline
& \fbtwo & \wnrr & \yg \\
\hline
\wdfivem~\cite{wd5m} & 13,603 & 2,414 & 57,542 \\
\dbpfivem (ours) & 13,030 & 2,129 & 114,352 \\
\wikidbp (ours) & 13,637 & 2,414 & 116,619 \\
\hline
\end{tabular}}}
\end{table}
 
By convention~\cite{TransE,RotatE}, 
we report the mean reciprocal rank ($MRR$) and $H@k$ $(k=1,10)$ results under the filter setting. 
Higher $MRR$ and $H@k$ indicate better performance.

\begin{table*}[!t]
\centering
\caption{\label{tab:link_prediction_wd5} Results on \fbtwo, \wnrr and \yg with \wdfivem as the background KG. 
The percentage in brackets indicates the result improvement compared with that in the LP setting.
The best results are marked in bold. 
}
\resizebox{1.0\textwidth}{!}{
\begin{tabular}{lllllllllll}
\hline
\multirow{2}{*}{Setting} &\multirow{2}{*}{Model} & \multicolumn{3}{c}{\fbtwo} & \multicolumn{3}{c}{\wnrr} & \multicolumn{3}{c}{\yg} \\
\cline{3-5}  \cline{6-8} \cline{9-11} & 
& $MRR$ & $H@10$ & $H@1$ 
& $MRR$ & $H@10$ & $H@1$ 
& $MRR$ & $H@10$ & $H@1$ \\ 
\hline
\multirow{7}{*}{LP}
& TransE~\cite{TransE} & 0.288 & 0.475 & $-$ & 0.224 & 0.510 & $-$ & 0.370 & 0.612 & 0.242 \\
& ConvE \cite{ConvE} & 0.325 & 0.501 & 0.237 & 0.430 & 0.520 & 0.400 & 0.440 & 0.620 & 0.350 \\
& RotatE \cite{RotatE} & 0.338 & 0.533 & 0.241 & \textbf{0.476} & \textbf{0.571} & 0.428 & 0.495 & 0.670 & 0.402 \\ 
& TuckER \cite{TuckER} & {0.358} & {0.544} & {0.266} & 0.470 & 0.526 & 0.443 & 0.505 & 0.661 & 0.422 \\
\cline{2-11}
& \modelnamernn & 0.285 & 0.433 & 0.210 & 0.406 & 0.447 & 0.382 & 0.447 & 0.617 & 0.354 \\
& \modelnamersn & 0.295 & 0.468 & 0.211 & 0.428 & 0.490 & 0.398 & 0.504 & 0.654 & 0.422 \\
& \modelnameformer & 0.307 & 0.481 & 0.220 & 0.444 & 0.494 & 0.417 & 0.523 & 0.678 & 0.439 \\
\hline
\multirow{7}{*}{JointLP}
& TransE & 0.346 (20.1\%) & 0.562 (18.3\%) & 0.237 ($-$) & 0.234 (4.4\%) & 0.528 (3.5\%) & 0.059 ($-$) & 0.504 (\textbf{36.2\%}) & 0.699 (14.2\%) & 0.294 (21.5\%) \\
& ConvE & 0.377 (16.0\%) & 0.575 (14.8\%) & 0.279 (17.7\%) & 0.455 (\textbf{5.8\%}) & 0.516 (-0.7\%) & 0.424 (\textbf{6.0\%}) & 0.467 (6.1\%) & 0.660 (6.5\%) & 0.387 (10.6\%) \\
& RotatE & 0.375 (10.9\%) & 0.579 (7.9\%) & 0.271 (12.4\%) & \textbf{0.476} (0.0\%)& 0.556 (-2.6\%) & 0.430 (4.7\%) & 0.502 (1.4\%) & 0.719 (7.3\%) & 0.425 (5.7\%) \\
& TuckER & 0.386 (7.8\%) & 0.579 (6.4\%) & 0.286 (7.5\%) & 0.463 (-1.5\%) & 0.518 (-1.5\%) & \textbf{0.432} (-2.5\%) & 0.542 (7.3\%) & 0.699 (5.7\%) & 0.459 (8.8\%) \\
\cline{2-11}
& \modelnamernn & 0.383 (34.4\%) & 0.581 (34.2\%) & 0.285 (35.7\%) & 0.398 (-2.0\%) & 0.475 (\textbf{6.3\%}) & 0.356 (-6.8\%) & 0.565 (26.4\%) & 0.739 (19.8\%) & 0.468 (32.2\%) \\ 
& \modelnamersn & {0.415} (40.7\%) & {0.612} (30.8\%) & {0.315} (49.3\%) & {0.427} (-0.2\%) & {0.499} (1.8\%) & 0.384 (-3.5\%) & 0.626 (24.2\%) & 0.791 (20.9\%) & 0.531 (25.8\%) \\
& \modelnameformer & 0.400 (30.3\%) & 0.592 (23.1\%) & 0.303 (37.7\%) & 0.426 (-4.1\%) & 0.496 (0.4\%) & {0.385} (-7.7\%) & {0.664} (27.0\%) & {0.806} (18.9\%) & {0.581} (32.3\%) \\
\hline
\multirow{3}{*}{PR4LP}
& \modelnamernn 
& 0.388 (36.1\%) & 0.586 (35.3\%) & 0.289 (37.6\%) 
& 0.418 (3.0\%) & 0.474 (6.0\%) & 0.385 (0.8\%) 
& 0.584 (30.6\%) & 0.748 (21.2\%) & 0.490 (38.4\%) \\
& \modelnamersn 
& 0.435 (\textbf{47.5\%}) & 0.644 (\textbf{37.6\%}) & 0.327 (\textbf{55.0\%}) 
& 0.430 (0.5\%) & 0.500 (2.0\%) & 0.388 (-2.5\%) 
& 0.656 (30.2\%) & 0.802 (\textbf{22.6\%}) & 0.565 (33.9\%) \\
& \modelnameformer 
& \textbf{0.440} (43.3\%) & \textbf{0.648} (34.7\%) & \textbf{0.336} (52.7\%) 
& 0.459 (3.4\%) & 0.522 (5.7\%) & 0.426 (2.2\%) 
& \textbf{0.687} (31.4\%) & \textbf{0.818} (20.6\%) & \textbf{0.608} (\textbf{38.5\%}) \\
\hline
\end{tabular}}
\end{table*}

\subsection{Results under Different Schemata}\label{sect:results}
The results with background KGs \wdfivem, \wikidbp and \dbpfivem are in Tables~\ref{tab:link_prediction_wd5}, \ref{tab:link_prediction_wd_dbp5} and \ref{tab:link_prediction_dbp5} (in Appendix~\ref{app:dbp5m_results}), respectively.

\myparagraph{LP}
From Table~\ref{tab:link_prediction_wd5}, we can see that our student models achieve comparable performance against existing baselines.
For example, \modelnameformer even achieves higher $H@1$ scores than RotatE on \yg.
Please note that our motivation is not to develop the best-performing link prediction model for LP.
Instead, we focus on the effective transferable embeddings in multi-source KGs.

\myparagraph{JointLP}
All models receive obvious improvement via joint representation learning.
The $MRR$ of TransE increases from $0.288$ to $0.346$.
This improvement demonstrates that joint representation learning is also able to transfer knowledge for helping target KG embeddings.
We see no improvement on \wnrr due to the limited common knowledge shared by the background KG and \wnrr.
Our variants outperform existing models by a large margin. 
This is because our path modeling can effectively capture the long-term relational dependency between KGs, which also benefits knowledge transfer.
The shortcoming of JointLP is the high time cost.

\begin{table*}[!t]
\centering
\caption{\label{tab:link_prediction_wd_dbp5} Results on \fbtwo, \wnrr and \yg with \wdfivem and \dbpfivem (\wikidbp) as background KGs.}
\resizebox{1.0\textwidth}{!}{
\begin{tabular}{lllllllllll}
\hline
\multirow{2}{*}{Setting} &\multirow{2}{*}{Model} & \multicolumn{3}{c}{\fbtwo} & \multicolumn{3}{c}{\wnrr} & \multicolumn{3}{c}{\yg} \\
\cline{3-5}  \cline{6-8} \cline{9-11} & 
& $MRR$ & $H@10$ & $H@1$ 
& $MRR$ & $H@10$ & $H@1$ 
& $MRR$ & $H@10$ & $H@1$ \\ 
\hline
\multirow{7}{*}{JointLP}
& TransE & 0.362 (25.7\%) & 0.574 (20.8\%) & 0.248 ($-$) & 0.235 (4.9\%) & 0.526 (3.1\%) & 0.060 ($-$) & 0.501 (35.4\%) & 0.703 (14.9\%) & 0.314 (29.8\%) \\
& ConvE & 0.375 (15.4\%) & 0.572 (14.2\%) & 0.270 (13.9\%) & 0.451 (4.9\%) & 0.513 (-1.3\%) & 0.419 (4.8\%) & 0.479 (8.9\%) & 0.681 (9.8\%) & 0.403 (14.8\%) \\
& RotatE & 0.384 (13.6\%) & 0.585 (9.8\%) & 0.276 (14.5\%) & \textbf{0.484} (1.7\%) & \textbf{0.558} (2.3\%) & 0.443 (3.5\%) & 0.516 (4.2\%) & 0.725 (8.2\%) & 0.439 (9.2\%) \\
& TuckER & 0.422 (17.9\%) & 0.605 (13.1\%) & 0.327 (22.9\%) & 0.471 (0.2\%) & 0.520 (-1.1\%) & \textbf{0.445} (0.5\%) & 0.553 (9.5\%) & 0.701 (6.1\%) & 0.473 (12.1\%) \\
\cline{2-11}
& \modelnamernn & 0.417 (46.3\%) & 0.620 (\textbf{43.2\%}) & 0.310 (47.6\%) & 0.400 (-1.5\%) & 0.476 (\textbf{6.5\%}) & 0.362 (-5.2\%) & 0.685 (53.2\%) & 0.851 (\textbf{37.9\%}) & 0.586 (65.5\%) \\
& \modelnamersn & 0.432 (46.4\%) & 0.625 (33.5\%) & 0.329 (55.9\%) & 0.412 (-3.7\%) & 0.487 (-0.6\%) & 0.372 (-6.5\%) & 0.693 (37.5\%) & 0.870 (33.0\%) & 0.584 (38.4\%) \\
& \modelnameformer & 0.437 (42.3\%) & 0.650 (35.1\%) & 0.335 (52.2\%) & 0.446 (0.5\%) & 0.504 (2.0\%) & 0.414 (-0.7\%) & {0.717} (37.1\%) & {0.878} (29.5\%) & {0.620} (41.2\%) \\
\hline
\multirow{3}{*}{PR4LP}
& \modelnamernn & 0.397 (39.3\%) & 0.600 (38.6\%) & 0.297 (41.4\%) & 0.421 (\textbf{5.2\%}) & 0.478 (0.4\%) & 0.388 (\textbf{7.2\%}) & 0.687 (\textbf{53.7\%}) & 0.841 (36.3\%) & 0.591 (\textbf{66.9\%}) \\
& \modelnamersn & 0.446 (\textbf{51.2\%}) & 0.653 (39.5\%) & 0.340 (\textbf{61.1\%}) & 0.407 (-4.9\%) & 0.488 (-0.4\%) & 0.365 (-6.5\%) & 0.693 (37.5\%) & 0.872 (33.3\%) & 0.594 (40.8\%) \\
& \modelnameformer & \textbf{0.454} (47.9\%) & \textbf{0.663} (37.8\%) & \textbf{0.348} (58.2\%) & 0.446 (0.5\%) & 0.509 (3.0\%) & 0.415 (-0.5\%) & \textbf{0.722} (38.0\%) & \textbf{0.880} (29.8\%) & \textbf{0.628} (43.1\%) \\
\hline
\end{tabular}}
\end{table*}

\myparagraph{PR4LP}
On \fbtwo and \yg, our variants receive a significant boost thanks to the effective knowledge transfer from background KGs.
We find that, on \wnrr, our framework does not bring obvious improvement.
We think there are two reasons.
First, \wnrr is an ontology-level KG, whereas the background KGs are instance-level.
Their facts have varying degrees of granularity, and have little common knowledge.
Second, as shown in Table~\ref{tab:links}, there are only thousands of entity alignment pairs to bridge the target and background KGs.
The limited size of entity alignment also hinders knowledge transfer. 
Our variants all outperform those in JointLP, which validates the effectiveness of our local re-training.

\begin{table*}[!t]
\centering
\caption{\label{tab:link_prediction_dbp} Results on DBP-ZH, DBP-JA and DBP-FR with DBP-EN as the background KG.}
\resizebox{0.999\textwidth}{!}{
\begin{tabular}{lllllllllll}
\hline
\multirow{2}{*}{Setting} &\multirow{2}{*}{Model} & \multicolumn{3}{c}{DBP-ZH} & \multicolumn{3}{c}{DBP-JA} & \multicolumn{3}{c}{DBP-FR} \\
\cline{3-5}  \cline{6-8} \cline{9-11} & 
& $MRR$ & $H@10$ & $H@1$ 
& $MRR$ & $H@10$ & $H@1$ 
& $MRR$ & $H@10$ & $H@1$ \\ 
\hline
\multirow{7}{*}{LP}
& TransE & 0.335 & 0.675 & 0.153 & 0.304 & 0.683 & 0.113 & 0.330 & 0.663 & 0.160 \\
& ConvE & 0.421 & 0.665 & 0.292 & 0.395 & 0.689 & 0.254 & 0.410 & 0.678 & 0.280 \\
& RotatE & 0.448 & 0.642 & 0.329 & 0.500 & 0.691 & 0.397 & 0.449 & 0.675 & 0.329\\
& TuckER & 0.455 & 0.625 & 0.360 & 0.483 & 0.665 & 0.392 & 0.461 & 0.648 & 0.366 \\
\cline{2-11}
& \modelnamernn & 0.394 & 0.594 & 0.287 & 0.405 & 0.568 & 0.318 & 0.387 & 0.554 & 0.299 \\
& \modelnamersn & 0.408 & 0.602 & 0.305 & 0.425 & 0.610 & 0.325 & 0.399 & 0.593 & 0.296 \\
& \modelnameformer & 0.450 & 0.643 & 0.345 & 0.482 & 0.667 & 0.385 & 0.447 & 0.649 & 0.339 \\ 
\hline
\multirow{7}{*}{JointLP}
& TransE & 0.418 (24.8\%) & 0.771 (14.2\%) & 0.214 (39.9\%) & 0.392 (28.9\%) & 0.775 (13.5\%) & 0.183 (61.9\%) & 0.417 (26.4\%) & 0.765 (15.4\%) & 0.227 (41.9\%) \\
& ConvE & 0.531 (26.1\%) & 0.775 (16.5\%) & 0.400 (37.0\%) & 0.517 (30.9\%) & 0.785 (13.9\%) & 0.376 (48.0\%) & 0.528 (28.8\%) & 0.763 (12.5\%) & 0.403 (43.9\%) \\
& RotatE & 0.539 (20.3\%) & 0.766 (19.3\%) & 0.410 (24.6\%) & 0.601 (20.2\%) & 0.800 (15.8\%) & 0.487 (22.7\%) & 0.545 (21.4\%) & 0.773 (14.5\%) & 0.416 (26.4\%) \\
& TuckER & 0.577 (26.8\%) & 0.759 (21.4\%) & 0.477 (32.5\%) & 0.637 (31.9\%) & 0.786 (18.2\%) & 0.557 (40.1\%) & 0.615 (33.4\%) & 0.771 (19.0\%) & 0.532 (45.4\%) \\
\cline{2-11}
& \modelnamernn & 0.555 (40.9\%) & 0.748 (25.9\%) & 0.448 (56.1\%) & 0.526 (29.9\%) & 0.726 (27.8\%) & 0.420 (32.1\%) & 0.599 (54.8\%) & 0.789 (\textbf{42.4\%}) & 0.496 (65.9\%) \\
& \modelnamersn & 0.608 (49.0\%) & 0.816 (\textbf{35.5\%}) & 0.490 (60.7\%) & 0.613 (44.2\%) & 0.781 (28.0\%) & 0.509 (56.6\%) & 0.621 (55.6\%) & 0.819 (38.1\%) & 0.507 (71.3\%) \\
& \modelnameformer & 0.628 (39.6\%) & \textbf{0.822} (27.8\%) & 0.517 (49.9\%) & 0.645 (33.8\%) & \textbf{0.838} (25.6\%) & 0.536 (39.2\%) & 0.667 (49.2\%) & 0.834 (28.5\%) & 0.571 (68.4\%) \\
\hline
\multirow{3}{*}{PR4LP}
& \modelnamernn & 0.571 (44.9\%) & 0.769 (29.5\%) & 0.463 (61.3\%) & 0.594 (46.7\%) & 0.777 (\textbf{36.8\%}) & 0.494 (55.3\%) & 0.603 (55.8\%) & 0.781 (41.0\%) & 0.505 (68.9\%) \\
& \modelnamersn & 0.621 (\textbf{52.2\%}) & 0.809 (34.4\%) & 0.513 (\textbf{68.2\%}) & 0.640 (\textbf{50.6\%}) & 0.806 (32.1\%) & 0.547 (\textbf{68.3\%}) & 0.639 (\textbf{60.2\%}) & 0.807 (36.1\%) & 0.543 (\textbf{83.4\%}) \\
& \modelnameformer & \textbf{0.640} (42.2\%) & 0.817 (27.1\%) & \textbf{0.536} (55.4\%) & \textbf{0.676} (40.2\%) & 0.834 (25.0\%) & \textbf{0.582} (51.2\%) & \textbf{0.672} (50.3\%) & \textbf{0.836} (28.8\%) & \textbf{0.580} (71.1\%) \\
\hline
\end{tabular}}
\end{table*}

\subsection{Results under Similar Schemata} \label{sect:similar_results}
Table~\ref{tab:link_prediction_dbp} displays the results on DBP15K-LP.
The background KG is \dbpfivem.
As DBP15K-LP contains similar multilingual schemata, we delete a larger number of background triplets that can lead to test data leakage.
For example, we detect $891$ relation alignment pairs between the ZH and EN DBpedia and delete $3,546$ triplets from the background KG.
An obvious observation is that all baselines and our variants gain more improvement than those under different schemata.
For example, TransE gets a $39.9\%$ improvement on $H@1$ in DBP-ZH under the PR4LP setting.
This is because the similar schemata reduce the difficulty of knowledge transfer between different KGs, and these KGs with similar schemata also share more complementary relational semantics.
By contrast, as shown in Tables~\ref{tab:link_prediction_wd5} and \ref{tab:link_prediction_wd_dbp5}, the H@1 improvement of TransE by knowledge transfer in heterogeneous KGs is no more than $29.8\%$.
The improvement of our model variants is still greater than these baselines under similar schemata, which further demonstrates the effectiveness of our pre-training and re-training framework for knowledge transfer.
This experiment shows that similar schemata can make knowledge transfer easier than heterogeneous schemata.

\subsection{Efficiency Evaluation}\label{sect:time}
\begin{figure}[!t]
\centering
\subfigure[Running time in LP and PR4LP. The background KG is \wdfivem.]
{\includegraphics[width=0.46\linewidth]{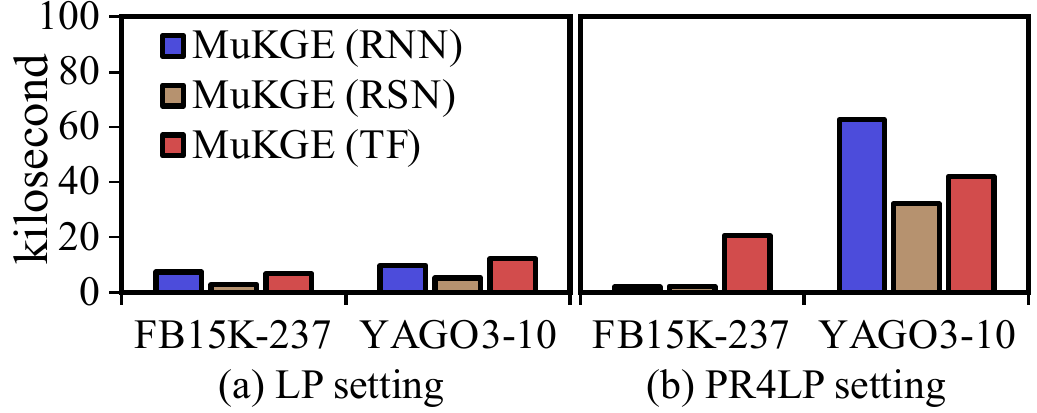}} 
\subfigure[Running time in the JointLP setting. The background KG is \wdfivem.]
{\includegraphics[width=0.46\linewidth]{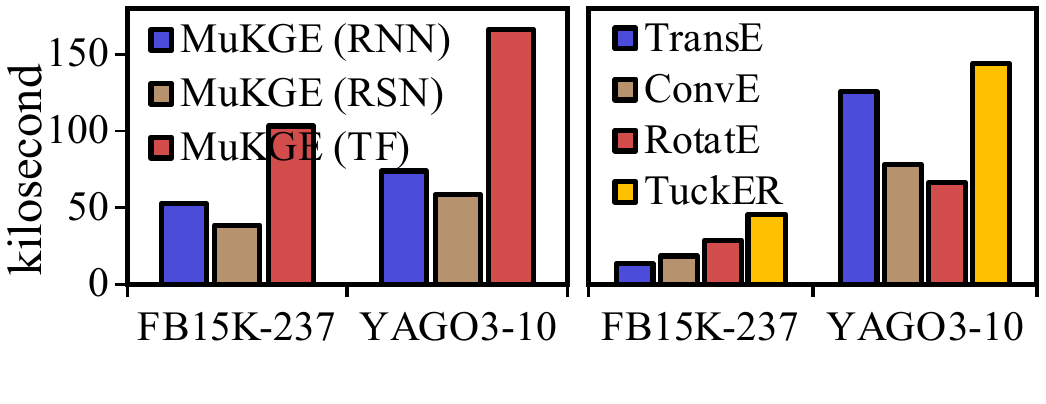}}
\caption{Running time (in kiloseconds) comparison. \label{fig:running_time}}
\end{figure}

Figure~\ref{fig:running_time} shows the training time.
In LP, \modelnameformer takes the most time, because its self-attention has a high computation complexity and requires lots of layers to obtain acceptable performance.
While in \modelnamernn and \modelnamersn, a two-layer RNN suffices.
\modelnamernn costs more time than \modelnamersn although they have a similar network architecture.
This is because the residual connections in RSN lead to fast convergence.
All our variants take more time on \yg than on \fbtwo due to the larger scale of \yg.
In PR4LP, as shown in the (b) part of Figure~\ref{fig:running_time} (a), \modelnameformer still costs more time than \modelnamersn.
As PR4LP introduces additional training resources (local subgraphs) and losses (knowledge distillation), the three variants all take more time than that in LP.
\modelnamersn achieves the highest efficiency in PR4LP.
In JointLP, our variants all cost more time than those in PR4LP and LP,
because learning background KGs greatly adds to training load.
We observe similar results in other datasets.

\subsection{RNN v.s. RSN v.s. Transformer} \label{sect:exp_encoder}
\modelnameformer obtains better performance than \modelnamernn and \modelnamersn in all settings due to the great representation expressiveness of Transformer \cite{Transformer}.
Generally, \modelnamersn outperforms \modelnamernn on most datasets. 
The relational residual connection in RSN can better capture relational semantics than LSTM. 
Note that \modelnamersn gets the most improvement in PR4LP.
For example, the H@1 score of \modelnamersn improves by $55.0\%$ compared with that in LP.
This is because RSN has fewer parameters than Transformer, which can ease the training of network knowledge distillation.
Although Transformer achieves slightly better results than RNN and RSN, 
it usually costs more training time, as shown in Figure~\ref{fig:running_time}.
Hence, we recommend RSN as the path encoder for a trade-off between effectiveness and efficiency.

\begin{table}[!t]
\centering
\caption{\label{tab:ablation_study} 
Ablation study in the PR4LP setting on \fbtwo and \yg with \wdfivem as the background KG. 
}
\resizebox{0.999\linewidth}{!}{\setlength\tabcolsep{15pt}
\begin{tabular}{lcccc}
\hline
\multirow{2}{*}{Model} & \multicolumn{2}{c}{\fbtwo} & \multicolumn{2}{c}{\yg} \\
\cline{2-3}  \cline{4-5} 
& $MRR$ & $H@1$ 
& $MRR$ & $H@1$  \\ 
\hline
\modelnamersn & 0.435 & 0.327 & 0.656 & 0.565 \\
\hline
\ \ w/o KD & 0.396 & 0.293 & 0.629 & 0.538 \\
\ \ w/o Feature KD & 0.418 & 0.313 & 0.629 & 0.534\\
\ \ w/o Network KD & 0.434 & 0.329 & 0.646 & 0.558 \\
\ \ w/o Prediction KD & 0.424 & 0.318 & 0.636 & 0.550 \\
\hline
\end{tabular}}
\end{table}

\subsection{Ablation Study} \label{sect:ablation}
Table~\ref{tab:ablation_study} presents the ablation study results on \modelnamersn.
We report the results of \modelnamernn and \modelnameformer in Table~\ref{tab:ablation_study_app} of Appendix~\ref{app:further_ablation}.
\modelname w/o KD is a special case of joint training with the linked subgraph.
Re-training augmented relational paths of the linked subgraph contributes the most to the performance improvement.
This shows that our subgraph sampling method can extract the most useful information to improve the target KG. 
We further observe that the proposed KD methods all bring additional improvements to \modelname w/o KD.
In general, the feature KD gives the greatest performance improvement among the three KD methods,
and the prediction KD comes second.
In re-training, the feature KD aligns the embedding space of the target KG with that of the linked subgraph, facilitating knowledge transfer from the background KGs to the target KG.
prediction KD enables the student model to predict with similar accuracy as the teacher model, which can also benefit the knowledge transfer.
Besides, the network KD can also deliver performance improvements.
This study validates the effectiveness of re-training with multi-level KD.

\subsection{Insufficiency of Joint Learning}\label{sect:joint_training}
\begin{figure}[!t]
\centering
\subfigure[In the naive JointLP setting.]
{\includegraphics[width=0.46\linewidth]{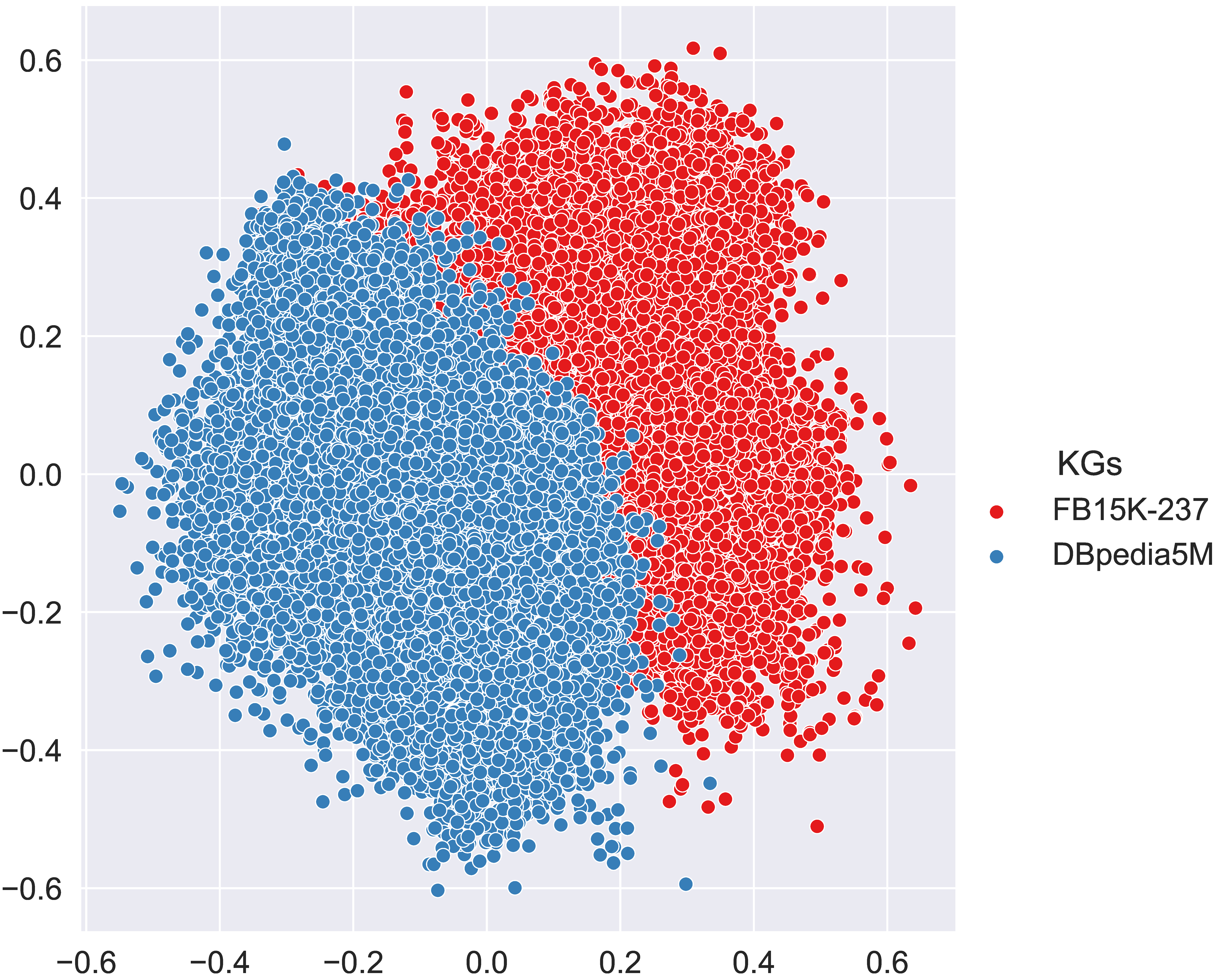}
\label{fig:dbp_fb_viz_a}} 
\subfigure[In the PR4LP setting.]
{\includegraphics[width=0.46\linewidth]{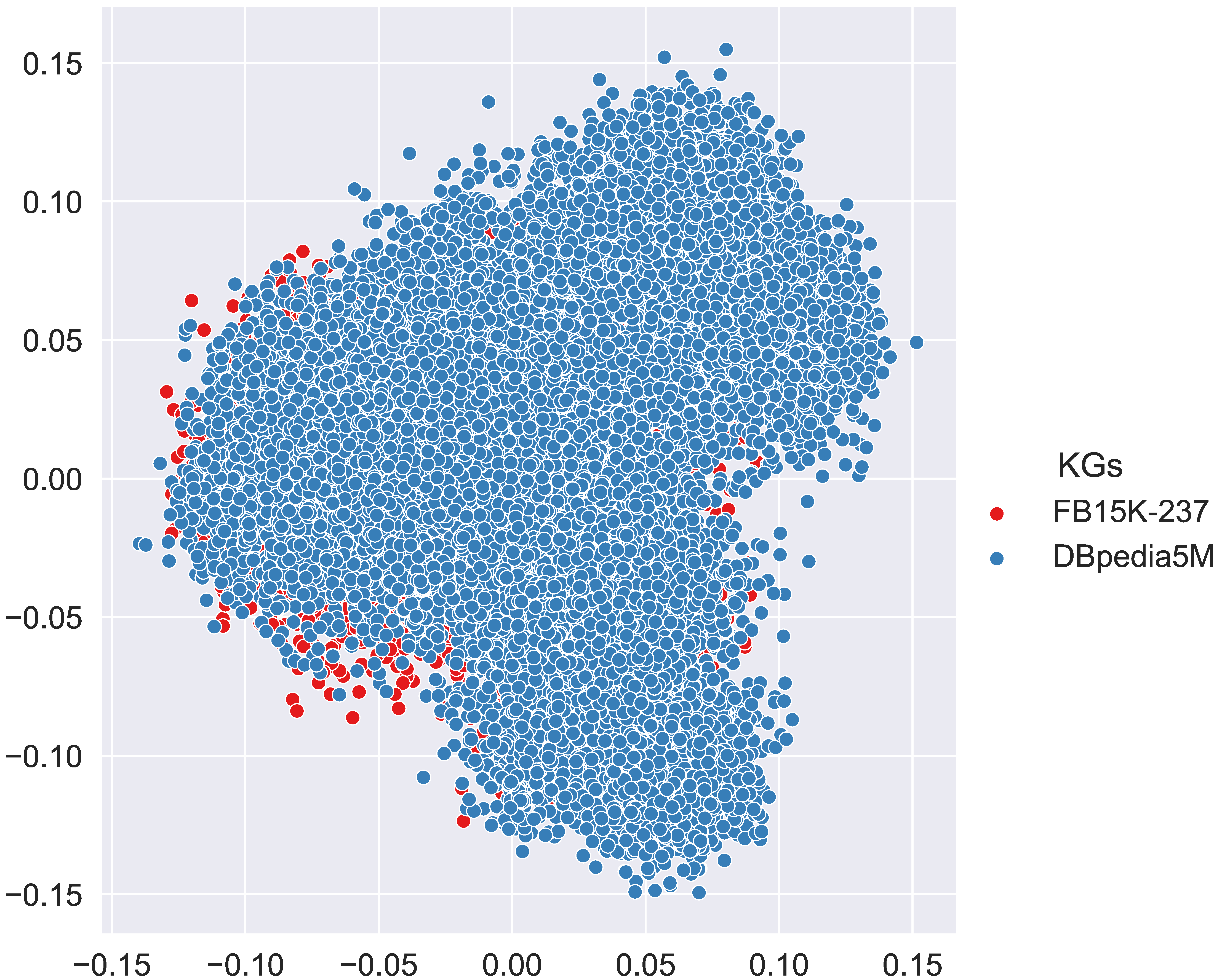}
\label{fig:dbp_fb_viz_b}}
\caption{Viz of entity embeddings of FB15K-237 entities and their counterparts in \dbpfivem.\label{fig:dbp_fb_viz}}
\end{figure}

We show in Figure~\ref{fig:dbp_fb_viz_a} and ~\ref{fig:dbp_fb_viz_b} the entity embeddings of \fbtwo entities and their counterparts in \dbpfivem, trained in the JointLP and PR4LP settings, respectively.
Although there is some overlap in the embedding distribution of \fbtwo entities and their counterparts, the majority of them are not aligned.
In this case, the knowledge in background KGs cannot be fully transferred into the target KG embeddings.
This may explain why joint learning fails to produce noticeable results.
This distribution is in distinct contrast to that in Figure~\ref{fig:dbp_fb_viz_b}, where embeddings are learned with PR4LP.
The embedding distributions of \fbtwo entities and their counterparts in the linked subgraph of \dbpfivem are closely aligned.
This indicates that our KD can align the embeddings of identical entities.
Such distribution alignment aids in effective knowledge transfer.
Overall, this experiment gives an insight into why our knowledge transfer via the linked subgraph outperforms joint learning.
See Appendix~\ref{app:viz} for another viz of embedding distribution.
More analyses on path length and convergence speed can be found in Appendix~\ref{app:further_ana}.

\subsection{Experiments on Multi-hop KGQA}\label{sect:exp_qa}

\begin{table}[!t]\setlength\tabcolsep{8pt}
\centering
\caption{\label{tab:qa} QA accuracy on \wqsp. Our model is pretrained with \wdfivem as the background KG.}
\resizebox{0.999\linewidth}{!}{
\begin{tabular}{lccc}
\hline
& Half-KG & Full-KG & Full-KG w/ rel. pruning \\
\hline
EmbedKGQA~\cite{MultiHopQA_KGE} & 0.485 & 0.587 & 0.666 \\
NSM~\cite{NSM_QA_WSDM21} & $-$ & $-$ & 0.743 \\
$\mu$KG~\cite{MuKG} & 0.547 & 0.646 & 0.723 \\  
\hline
EmbedKGQA + \modelname & 0.518 & 0.632  & 0.746  \\
\hline
\end{tabular}}
\end{table}

We follow EmbedKGQA~\cite{MultiHopQA_KGE}, the first embedding-based model for multi-hop KGQA, in this experiment.
EmbedKGQA learns a mapping between the embeddings of a question and its answer entity.
We improve EmbedKGQA by using our pre-training and re-training framework to learn KG embeddings.
We choose recent embedding-based methods NSM~\cite{NSM_QA_WSDM21} and $\mu$KG \cite{MuKG} as baselines.
See Appendix~\ref{app:qa} for the setup details.
Table~\ref{tab:qa} presents the accuracy results on \wqsp \cite{WebQuestionsSP}.
Our QA method, denoted by EmbedKGQA + \modelname, outperforms EmbedKGQA.
This is because our method can leverage the knowledge transfer from background KGs to produce improved embeddings for QA.
The results in the Full-KG setting are higher than those in the Half-KG setting because the KG incompleteness issue can reduce the expressiveness of KG embeddings.
This indicates the importance of our knowledge transfer in improving KG incompleteness.
Our method outperforms NSM in the Full-KG w/ rel. pruning setting.
NSM introduces a more complex reasoning method for KGQA, 
whereas our method transfers useful knowledge to improve the simple method EmbedKGQA with more expressive KG embeddings that can benefit KG-related downstream tasks.
$\mu$KG is a strong baseline with knowledge transfer, and our method outperforms it in the relation pruning setting.

\subsection{Case Study}\label{sect:case}

\begin{table}[!t]
\centering
\caption{\label{tab:rule_yg} Examples of the 1-hop and 2-hop rules mined from the joint graph of YAGO3 (YG) and Wikidata (WD).}
\resizebox{0.999\linewidth}{!}{\Large
\begin{tabular}{cccc}
\hline
Rule head & & Rule body & Conf. \\
\hline
\textit{YG\,:\,founder}\,(X, Y) & $\Longleftarrow$ & \textit{WD\,:\,foundedBy}\,(X, Y) & 0.84 \\
\hline
\textit{YG\,:\,parentOrganization}\,(X, Y) & $\Longleftarrow$ & \textit{WD\,:\,subsidiary}\,(Y, X) & 0.85 \\
\hline
\textit{YG\,:\,musicBy}\,(X, Y) & $\Longleftarrow$ & \textit{WD\,:\,composer}\,(X, Y) & 0.97 \\
\hline
\textit{YG\,:\,byArtist}\,(X, Y) & $\Longleftarrow$ & \textit{WD\,:\,performer}\,(X, Y) & 0.99 \\
\hline
\textit{YG\,:\,spouse}\,(X, Y) & $\Longleftarrow$ & \textit{WD\,:\,father}\,(Z, Y) $\wedge$ \textit{WD\,:\,mother}\,(Z, X) & 0.82 \\
\hline
\textit{YG\,:\,author}\,(X, Y) & $\Longleftarrow$ & \textit{WD\,:\,followedBy}\,(Z, X) $\wedge$ \textit{WD\,:\,notableWork}\,(Y, Z) & 0.84 \\
\hline
\textit{YG\,:\,byArtist}\,(X, Y) & $\Longleftarrow$ & \textit{WD\,:\,follows}\,(X, Z) $\wedge$ \textit{WD\,:\,performer}\,(Z, Y) & 0.97 \\
\hline
\textit{YG\,:\,partOfSeries}\,(X, Y) & $\Longleftarrow$ & \textit{WD\,:\,followedBy}\,(X, Z) $\wedge$ \textit{WD\,:\,partOfTheSeries}\,(Z, Y) & 0.98 \\
\hline
\end{tabular}}
\end{table}

Our framework succeeds on \fbtwo, \yg and DBP15K-LP, but fails on \wnrr.
In this section, we investigate the causes of success and failure from a rule-mining perspective.
Our assumption is that, if there exists transferable knowledge between the target and background KGs,
such knowledge can be represented by logic rules \cite{KGERule}.
We first merge the background and target KGs into a joint graph by merging aligned entities into a unified node.
Then, we use AMIE \cite{AMIE} to mine Horn rules from the joint graph.
Table~\ref{tab:rule_yg} lists some examples of the 1-hop and 2-hop rules.
We focus on the rules whose head and body cross two KGs since they indicate the potential knowledge that can be transferred. 
Specifically, the relations in the rule body are from \wdfivem, while the relation in the rule head is from \yg.
We totally get $105$ 1-hop and $58$ 2-hop rules across \wdfivem and \yg. 
We use a namespace like \textit{YG:} or \textit{WD:} to denote where the relation comes from.
We can see that the 1-hop rules capture the equivalent relations, implying that Wikidata could share some common relational facts with \yg, which leads to easy knowledge transfer.
Please note that we have deleted the majority of these triplets from background KGs because they could cause test data leakage.
2-hop rules capture the reasoning paths in Wikidata to represent a relation in \yg.
These 2-hop rules indicate that there exists complex and underlying knowledge could be transferred from Wikidata to \yg.

\begin{table}[!t]
\centering
\caption{\label{tab:rule_wn} Rule examples of \wnrr (WN) and WD.}
\resizebox{0.999\linewidth}{!}{\Large
\begin{tabular}{cccc}
\hline
Rule head & & Rule body & Conf. \\
\hline
\textit{WN\,:\,memberOfDomainRegion}\,(X, Y) & $\Longleftarrow$ & \textit{WD\,:\,countryOf Origin}\,(X, Y) & 1.00 \\
\hline
\textit{WN\,:\,hasPart}\,(X, Y) & $\Longleftarrow$ & \textit{WD\,:\,headquartersLocation}\,(Y, Z) $\wedge$ \textit{WD\,:\,ownedBy}\,(Z, X) & 0.80 \\
\hline
\textit{WN\,:\,hasPart}\,(X, Y) & $\Longleftarrow$ & \textit{WD\,:\,executiveBody}\,(Y, Z) $\wedge$ \textit{WD\,:\,country}\,(Z, X) & 0.63 \\
\hline
\textit{WN\,:\,memberMeronym}\,(X, Y) & $\Longleftarrow$ & \textit{WD\,:\,constellation}\,(Z, X) $\wedge$ \textit{WD\,:\,partOf}\,(Z, Y) & 1.00 \\
\hline
\textit{WN\,:\,synsetDomainTopicOf}\,(X, Y) & $\Longleftarrow$ & \textit{WD\,:\,narrativeLocation}\,(Z, X) $\wedge$ \textit{WD\,:\,presentInWork}\,(Z, Y) & 1.00 \\
\hline
\end{tabular}}
\end{table}

Table~\ref{tab:rule_wn} lists the rule examples mined from \wnrr and \wdfivem.
We totally get $1$ 1-hop and $86$ 2-hop rules.
Although the number of 2-hop rules is larger than that of rules across \wdfivem and \yg, the number of rule groundings from \wnrr and \wdfivem is much smaller than that of \wdfivem and \yg.
This means that \wdfivem and \wnrr only share very limited transferable knowledge.
Besides, the rule heads only relate to four relations in \wnrr, i.e., \textit{WordNet:\, memberOfDomainRegion}, \textit{WordNet: \,hasPart}, \textit{WordNet: \,memberMeronym}, and \textit{WordNet: \,synsetDomainTopicOf}.
The other $14$ relations in \wnrr have no counterparts or related reasoning paths in \wdfivem.
We think this finding can explain why our framework fails on \wnrr.
The main reason lies in the very limited transferable knowledge between the background KGs (e.g., \wdfivem) and \wnrr.

\section{Related Work and Discussion}\label{sect:related_work}

\myparagraph{Knowledge graph embedding.}\label{sect:related_work_kge}
We divide existing KG embedding models into three groups according to their technical characteristics. 
Translational techniques interpret a relation as the translation vector from its subject entity embedding to its object \cite{TransE,TransH,TransR}. 
Factorization-based models encode a KG as a binary tensor and decompose it into a set of matrices to represent entities and relations \cite{BilinearKGE,DistMult,SimplE,TuckER}.
Neural models embed KGs by doing the entity or relation prediction task through various neural architectures \cite{ProjE,ConvE,RGCN,CompGCN,CoKE,HittER}.
Interested readers can refer to the survey \cite{TKDE_KGE} and experimental studies \cite{lp_eval_sigmod,lp_eval_tkdd}.
These studies focus on learning on a single KG.
In contrast, our work seeks to enable effective knowledge transfer between multiple KGs.
In this work, we consider neural networks \cite{LSTM,RSN,Transformer} as encoders due to their great effectiveness and good expansibility to downstream models \cite{KG_survey}.

\myparagraph{Embedding-based entity alignment}\label{sect:related_work_ea}
Existing entity alignment methods mainly focus on capturing entity similarities.
In addition to an embedding learning model (e.g., TransE \cite{TransE} or GCN \cite{GCN}),
it has an alignment learning model to push semantically similar entities to have similar embeddings. 
Current studies improve entity alignment in two aspects.
The first seeks to enhance embedding learning by employing advanced representation learning techniques \cite{MuGNN,AliNet,HyperKA},
or incorporate more side features \cite{RDGCN,EMGCN,KDCoE,AttrE,MultiKE,RL4EA}.
The latter learns from seed entity alignment to capture the alignment relationships between different KGs. 
Interested readers can refer to recent surveys \cite{OpenEA,TKDE_EA_survey,OpenAI_EA_survey,VLDBJ_EA_survey}.
In our work, we only consider relational structures because they are the basic features of KGs.

\myparagraph{Pre-training and fine-tuning}
``Pre-training and fine-tuning'' has become the learning paradigm for NLP~\cite{NLP_pretraining_survey} and CV \cite{ImageDA_survey}.
However, it cannot be applied to multi-source KGs.
In NLP, the tokens with the same symbolic form refer to the same word.
This nature eliminates the need to consider data heterogeneity when training a large-scale text corpus.
The pre-trained and task-specific models can share the same vocabulary. 
The pixels and images in CV also have this nature.
The same values of RGB refer to the same color. 
In contrast, data heterogeneity is the central challenge for our setting.
A recent line of related work is incorporating KGs into language models \cite{ERNIE,K-Adapter,SKILL,KEPLMs_survey}.
In contrast, our work seeks knowledge transfer between KGs. 
We do not use language models.

Our work also differs from graph or GNN pre-training. 
Graph pre-training seeks to learn node embeddings or GNN encoders on a given graph without using task labels \cite{GraphPretraingSurvey2022}.
The pretrained embeddings or encoders are then used by the task.
Rather than focusing solely on pre-training embeddings on the target KG, our work investigates knowledge transfer between multiple data sources.

\section{Conclusion and Future Work}\label{sec:concl}
We present a novel framework, i.e., pre-training and re-training, for knowledge transfer between multi-source KGs. 
It first pre-trains a large teacher model over linked multi-source KGs,
and then re-trains a student model over small task-specific KGs by distilling knowledge from the teacher model.
We conduct comprehensive experiments on the ability of our pre-training and re-training framework for KG embeddings and application.
Future work may consider the side information such as entity descriptions and large language models to enhance KG pre-training.

\begin{acks}
This work is supported by the National Natural Science Foundation of China (No. 62272219) and the Alibaba Group through Alibaba Research Fellowship Program.
\end{acks}

\bibliographystyle{ACM-Reference-Format}
\bibliography{reference}

\clearpage
\appendix

\section{LP Dataset Statistics}\label{app:lp_dataset}
The statistics of the target link prediction datasets \fbtwo \cite{TransE}, \wnrr \cite{ConvE} and \yg \cite{YAGO3-10} are shown in Table~\ref{table:datasets}.

\begin{table}[!ht]
\centering
\caption{\label{table:datasets}Statistics of the target link prediction datasets.}
\resizebox{0.999\linewidth}{!}{
\begin{tabular}{lrrrrr}
\hline
\multirow{2}{*}{Dataset} & 
\multicolumn{1}{c}{\multirow{2}{*}{\# Entities}} & 
\multicolumn{1}{c}{\multirow{2}{*}{\# Relations}} &
\multicolumn{3}{c}{\# Triplets}\\
\cline{4-6}
& & 
& \multicolumn{1}{c}{\# Training}
& \multicolumn{1}{c}{\# Validation}
& \multicolumn{1}{c}{\# Test} \\
\hline
\fbtwo~\cite{TransE} & 14,541 & 237 & 272,115 & 17,535 & 20,466 \\
\wnrr~\cite{ConvE} & 40,943 & 11 & 86,835 & 3,034 & 3,134 \\
\yg~\cite{YAGO3-10} & 123,182 & 37 & 1,079,040 & 5,000 & 5,000 \\
\hline
\end{tabular}}
\end{table}

\section{Dataset Statistics of DBP15K-LP}\label{app:dbp_lp}
\begin{figure}[!ht]
\centering
\includegraphics[width=0.9\linewidth]{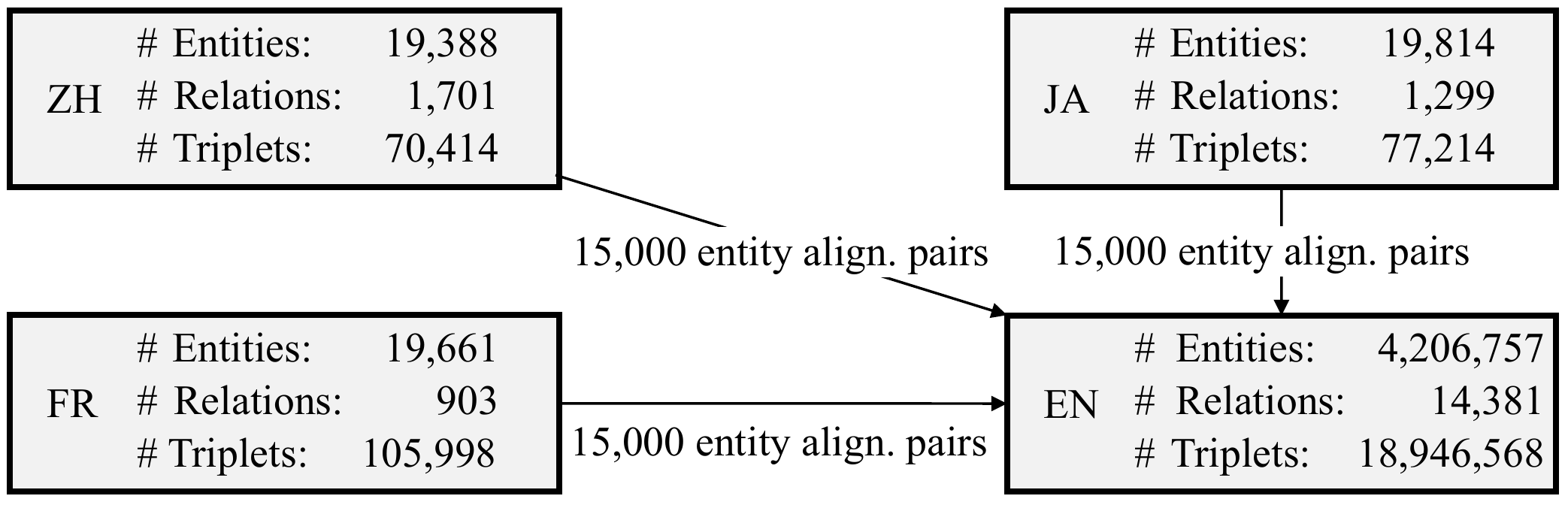}
\caption{Illustration of our \dbplp dataset for link prediction over multiple linked KGs with similar schemas.}
\label{fig:dbp15k_dataset_graph}
\end{figure}

The statistics of our proposed dataset DBP15K-LP are shown in Figure~\ref{fig:dbp15k_dataset_graph}.
It has four KGs and the ZH, JA, FR KGs have $15,000$ entity alignment pairs with the EN KG.

\section{Implementation Details} \label{app:parameter}
In the filter setting, given a test query, if a candidate completes the test query and the resulting triplet appears in training data, we remove this entity from the candidate list because the model has already seen the triplet.
We use Xavier \cite{Xavier_init} to initialize trainable parameters.
We use Adam \cite{Adam} to optimize the loss,
and dropout \cite{Dropout} to help prevent overfitting.
We select the embedding dimension from $\{128,256,512,768,1024\}$,
path length from $\{3,5,7,9,11\}$,
learning rate from $\{0.00001, 0.0001, 0.0002, 0.0005, 0.001, 0.002, 0.005, 0.01\}$,
the number of layers from $\{1,2,4,8,12\}$,
batch size from $\{1024, 2048, \\ 4096, 8192, 10240, 20480, 40960\}$,
dropout rate from $\{0.1, 0.2, 0.3, 0.4,\\ 0.5\}$,
$\alpha$ and $\beta$ in $\{0.1, 0.2, 0.3, ..., 0.9\}$.
The number of random walks starting from each entity is $2$.
We did not enumerate and try all possible combinations of hyper-parameter settings due to the massive training time over the huge data scale,
Hence, some hyper-parameters such as the batch size were set empirically or according to our empirical experiments on small data.

\section{LP Results with DBpedia5M}\label{app:dbp5m_results}
The link prediction results under the DS setting with DBpedia5M as the background KG are given in Table~\ref{tab:link_prediction_dbp5}.

\begin{table*}[!t]
\centering
\caption{\label{tab:link_prediction_dbp5} Results on \fbtwo, \wnrr and \yg with \dbpfivem as the background KG.}
\resizebox{1.0\textwidth}{!}{
\begin{tabular}{lllllllllll}
\hline
\multirow{2}{*}{Setting} &\multirow{2}{*}{Model} & \multicolumn{3}{c}{\fbtwo} & \multicolumn{3}{c}{\wnrr} & \multicolumn{3}{c}{\yg} \\
\cline{3-5}  \cline{6-8} \cline{9-11} & 
& $MRR$ & $H@10$ & $H@1$ 
& $MRR$ & $H@10$ & $H@1$ 
& $MRR$ & $H@10$ & $H@1$ \\ 
\hline
\multirow{5}{*}{JointLP}
& TransE & 0.320 (11.1\%) & 0.547 (15.2\%) & 0.206 ($-$) & 0.231 (3.1\%) & 0.524 (2.7\%) & 0.057 ($-$) & 0.482 (30.2\%) & 0.686 (12.1\%) & 0.289 (19.4\%) \\
& ConvE & 0.360 (10.8\%) & 0.553 (10.4\%) & 0.263 (11.0\%) & 0.443 (3.0\%) & 0.510 (-1.9\%) & 0.407 (1.8\%) & 0.463 (5.2\%) & 0.667 (7.6\%) & 0.382 (9.1\%) \\
& RotatE & 0.379 (12.1\%) & 0.581 (3.8\%) & 0.275 (14.1\%) & {0.474} (-0.4\%) & \textbf{0.553} (-3.1\%) & \textbf{0.431} (0.7\%) & 0.482 (-2.6\%) & 0.691 (3.1\%) & 0.431 (7.2\%) \\
& TuckER & 0.382 (6.7\%) & {0.588} (8.1\%) & 0.281 (5.6\%) & 0.458 (-2.5\%) & 0.512 (-2.7\%) & 0.427 (-3.6\%) & 0.535 (5.9\%) & 0.702 (6.2\%) & 0.464 (10.0\%) \\
\cline{2-11}
& \modelnamernn & 0.362 (27.0\%) & 0.538 (24.2\%) & 0.272 (29.5\%) & 0.374 (-7.9\%) & 0.443 (-0.9\%) & 0.333 (-12.8\%) & 0.673 (50.6\%) & 0.843 (\textbf{36.6\%}) & 0.572 (61.6\%) \\
& \modelnamersn & {0.396} (34.2\%) & 0.580 (23.9\%) & 0.304 (40.1\%) & 0.406 (-5.1\%) & 0.480 (-2.0\%) & 0.361 (-9.2\%) & 0.687 (36.3\%) & 0.866 (32.4\%) & 0.579 (37.2\%) \\
& \modelnameformer & {0.396} (29.0\%) & 0.579 (20.3\%) & {0.305} (38.6\%) & 0.449 (1.1\%) & 0.500 (1.2\%) & 0.419 (0.5\%) & {0.713} (36.3\%) & {0.871} (28.5\%) & {0.616} (40.3\%) \\
\hline
\multirow{3}{*}{PR4LP}
& \modelnamernn & 0.378 (32.6\%) & 0.563 (30.0\%) & 0.286 (36.2\%) & 0.372 (-7.9\%) & 0.442 (-1.1\%) & 0.335 (-12.3\%) & 0.679 (\textbf{51.9\%}) & 0.835 (35.3\%) & 0.583 (\textbf{64.7\%}) \\
& \modelnamersn & 0.408 (\textbf{38.3\%}) & 0.611 (\textbf{30.6\%}) & 0.308 (\textbf{46.0\%}) & 0.407 (-4.7\%) & 0.490 (0.0\%) & 0.362 (-9.0\%) & 0.695 (38.0\%) & 0.870 (33.0\%) & 0.586 (38.9\%) \\
& \modelnameformer & \textbf{0.424} (38.1\%) & \textbf{0.622} (29.3\%) & \textbf{0.315} (43.1\%) & \textbf{0.458} (\textbf{3.2\%}) & {0.511} (\textbf{3.4\%}) & {0.429} (\textbf{2.9\%}) & \textbf{0.718} (37.3\%) & \textbf{0.881} (29.9\%) & \textbf{0.621} (41.5\%) \\
\hline
\end{tabular}}
\end{table*}

\section{Entity Embedding Distribution}\label{app:viz}

\begin{figure}[!ht]
\centering
\includegraphics[width=0.8\linewidth]{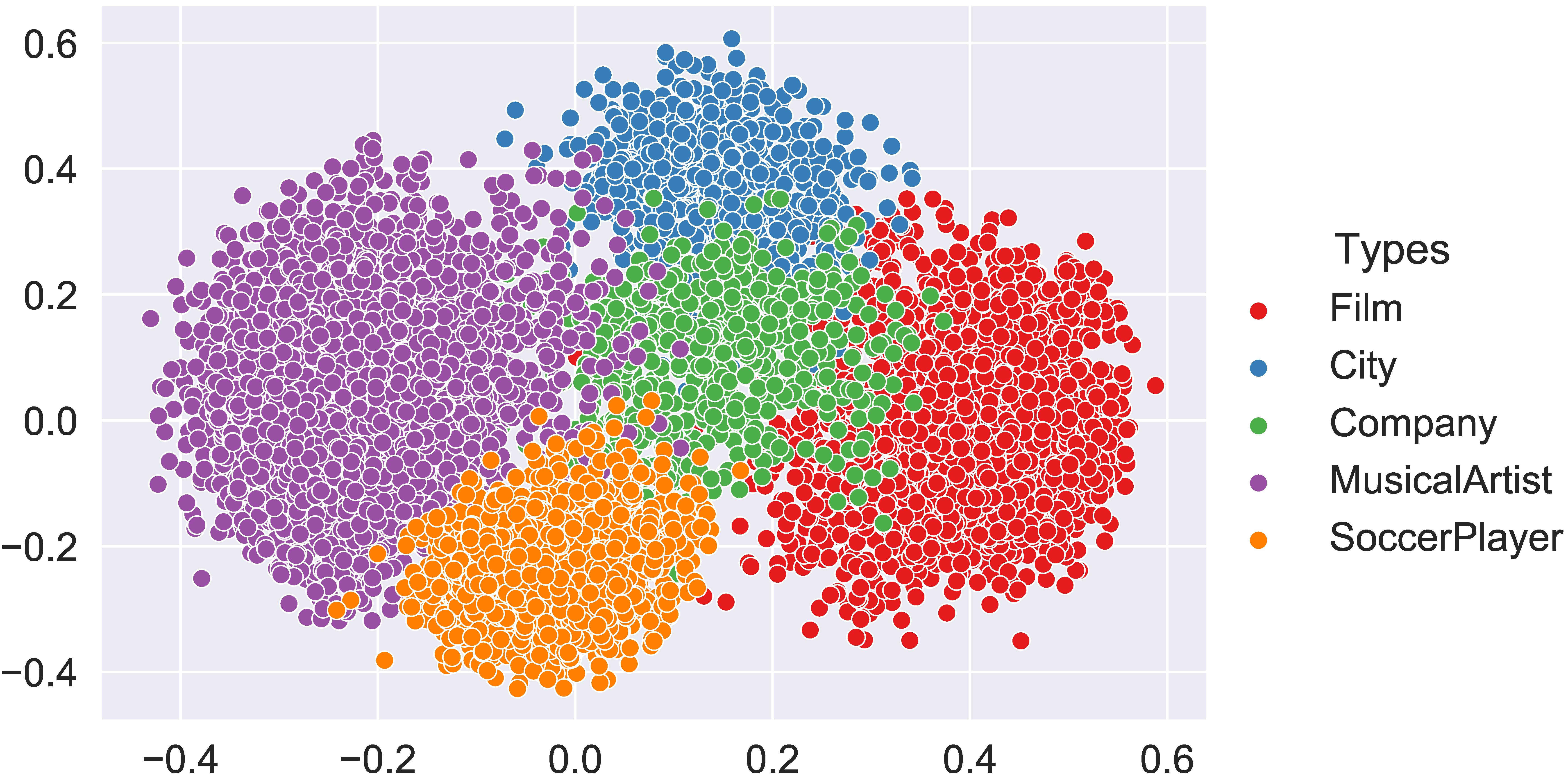}
\caption{Visualization of the entity embeddings of five popular types in \dbpfivem.}
\label{fig:dbp_viz}
\end{figure}

Figure~\ref{fig:dbp_viz} shows a visualization of the projected 2-dimensional entity embeddings of five popular types ``Film'', ``City'', ``Company'', ``Musical Artist'', and ``Soccer Player'' in \dbpfivem.
The dimension reduction method is principal component analysis.
We can find that similar entities are clustered while the entities of different types are generally separated in the embedding space, 
which shows that our encoder is able to capture entity similarities.

\section{Further Analyses of LP}\label{app:further_ana}
In this section, we present the results and analyses of link prediction.

\subsection{Ablation Study on \modelnamernn and (TF)}\label{app:further_ablation}
The results of our ablation study on \modelnamernn and \modelnameformer are given in Table~\ref{tab:ablation_study_app}.
We can observe similar results and come to the same conclusion as discussed in Sect.~\ref{sect:ablation}.

\begin{table}[!ht]
\centering
\caption{\label{tab:ablation_study_app} 
Ablation study in the PR4LP setting on \fbtwo and \yg with \wdfivem as the background KG. }
\resizebox{0.999\linewidth}{!}{\setlength\tabcolsep{14pt}
\begin{tabular}{lcccc}
\hline
\multirow{2}{*}{Method} & \multicolumn{2}{c}{\fbtwo} & \multicolumn{2}{c}{\yg} \\
\cline{2-3}  \cline{4-5} 
& $MRR$ & $H@1$ 
& $MRR$ & $H@1$  \\ 
\hline
\modelnamernn & 0.388 & 0.289 & 0.584 & 0.490 \\
\hline
\ \ w/o KD & 0.370 & 0.278 & 0.558 & 0.459 \\
\ \ w/o Feature KD & 0.379 & 0.284 & 0.567 & 0.466 \\
\ \ w/o Network KD & 0.382 & 0.287 & 0.580 & 0.485 \\
\ \ w/o Prediction KD & 0.379 & 0.285 & 0.568 & 0.468 \\
\hline
\modelnameformer & 0.440 & 0.336 & 0.687 & 0.608 \\
\midrule
\ \ w/o KD & 0.413 & 0.320 & 0.658 & 0.572 \\
\ \ w/o Feature KD & 0.425 & 0.321 & 0.664 & 0.581 \\
\ \ w/o Network KD & 0.433 & 0.334 & 0.684 & 0.607 \\
\ \ w/o Prediction KD & 0.430 & 0.329 & 0.677 & 0.592 \\
\hline
\end{tabular}}
\end{table}

\subsection{Performance w.r.t Path Length} \label{app:further_ana_length}
\begin{figure}[!ht]
\centering
\includegraphics[width=0.999\linewidth]{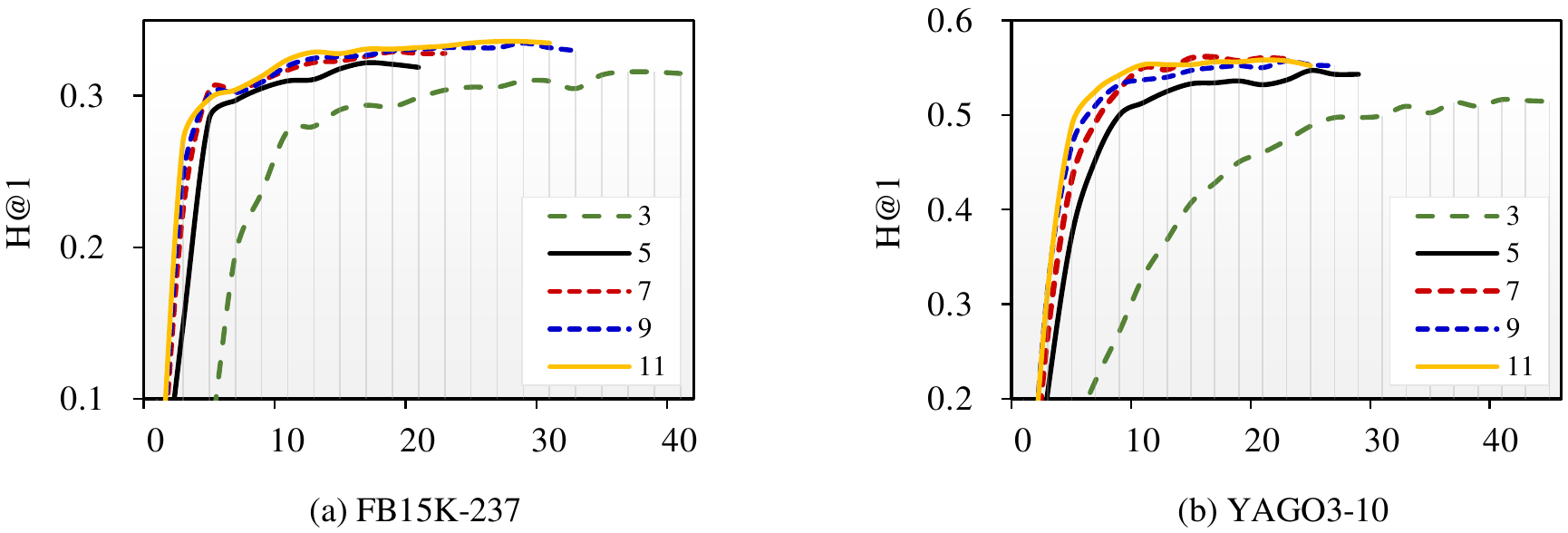}
\caption{$H@1$ results on the validation data evaluated every two epochs w.r.t the path length in the PR4LP setting, where the background KG is \wdfivem and the encoder is RSN.}
\label{fig:path_len}
\end{figure}

We study the effect of different path lengths on link prediction.
Figure~\ref{fig:path_len} shows the $H@1$ results validated every two epochs in the training process of \modelnamersn.
The background KG is \wdfivem.
The findings are summarized as follows.
First, learning with relational paths of length $3$ (the path is degraded into a triplet) converges slower and performs worse than with longer paths.
The reason is that longer paths have higher efficacy in capturing both relational semantics in a single KG and those across different KGs.
Second, in general, learning with longer paths leads to better performance, but it costs more training time for each epoch.
Finally, learning with paths of length $5$ or $7$ could achieve similar performance to that with longer paths.
For a trade-off between effectiveness and efficiency, we use paths of length $5$.

\subsection{Convergence Speed Comparison} \label{app:further_ana_convergence}
\begin{figure}[!ht]
\centering
\includegraphics[width=0.999\linewidth]{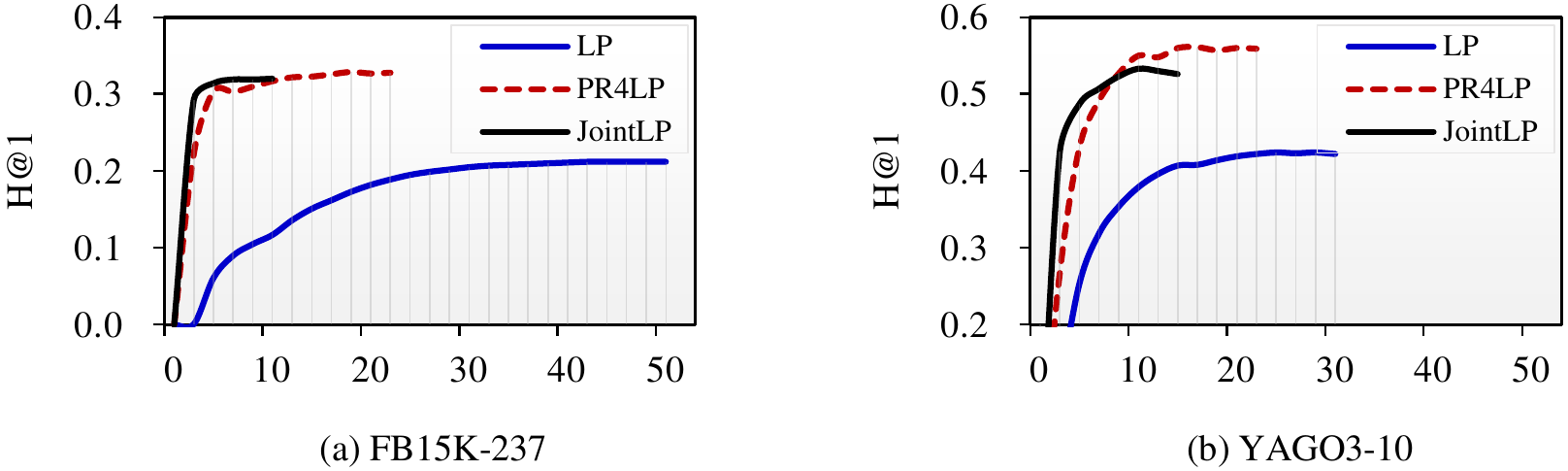}
\caption{$H@1$ performance validated every two epochs of \modelname (RSN) in the LP and KT4LP settings.}
\label{fig:hits1_epochs}
\end{figure}

Figure~\ref{fig:hits1_epochs} compares the convergence speed of \modelname in the LP, JointLP and PR4LP settings, where the encoder is RSN and the background KG is \wdfivem.
Thanks to the background knowledge from Wikidata, \modelname converges faster in the JointLP and PR4LP settings than in the LP setting.
\modelname takes fewer training epochs in JointLP than LP and PR4LP.
Because of the large scale training data in JointLP, it is easily overfit.

\section{Setup for Multi-hop KGQA}\label{app:qa}
In this section, we report the experimental results on the auxiliary task, multi-hop KGQA, as an application of the proposed re-training and pre-training framework.
The task seeks to retrieve an entity from a target KG as the answer to a natural language question, which may require multi-hop relational reasoning over the KG.
For example, the answer to ``What are the genres of movies written by Louis Mellis?'' is ``Crime''.
It needs two-hop reasoning to answer the question.
The first is to obtain the movies written by Louis Mellis, and the second is to get the genres of these movies.
The reasoning path in a KG could be 
$\textit{Louis Melli} \xrightarrow[]{\textit{write}} \textit{X} \xrightarrow[]{\textit{genre}} \text{?}$.

Conventional semantic parsing-based QA methods convert the question to a logical form (e.g., a SPARQL query) 
and execute it over the KG to search for an answer \cite{CaseKGQA}.
They suffer from the parsing errors and KG incompleteness.
Recent embedding-based methods \cite{MultiHopQA_KGE,NSM_QA_WSDM21} learn an embedding similarity of a question-answer pair.
KG embeddings can capture some semantics of the missing triplets in KGs, which could help resolve the KG incompleteness issue and benefit KGQA.
In this experiment, we would like to show that, through knowledge transfer, our framework could further improve KG embeddings and thus boost the final performance of KGQA.

\myparagraph{Settings}
We follow EmbedKGQA \cite{MultiHopQA_KGE} to develop our QA method.
EmbedKGQA is the first embedding-based method for multi-hop KGQA.
Its key idea is to learn a mapping between the representation of a question and the embedding of its answer entity.
EmbedKGQA learns embeddings for the target KG using TransE \cite{TransE} or ComplEx~\cite{ComplEx}.
Our \modelname could also be used here to learn embeddings with the knowledge transferred from other background KGs.
EmbedKGQA uses RoBERTa \cite{RoBERTa} to represent natural questions. 
In our QA method,
we use our pre-training and re-training framework to learn transferable embeddings for the target KG with the help of the background KG \wdfivem.
To ensure a fair comparison, other QA modules remain the same as those in EmbedKGQA.

\myparagraph{Dataset}
In our experiment, we chose the widely used multi-hop QA dataset \wqsp \cite{WebQuestionsSP} following EmbedKGQA.
The dataset uses Freebase entities \cite{Freebase} to answer multi-hop questions.
The number of questions is $4,737$.
Following EmbedKGQA and $\mu$KG, we use the extracted subset of Freebase as the target KG.
It contains the relation triplets that are within the two hops of the entities specified in WebQuestionsSP questions.
It also only includes the relations specified in the questions.
The resulting KG, denoted as FB4QA, for answering these questions has about $1.8$ million entities and $5.7$ million triplets in total.
To learn embeddings for FB4QA using our framework, we use \wdfivem as the background KG.
It has $493,987$ entity alignment pairs to FB4QA, which can be used for knowledge transfer via the proposed retraining method.

\myparagraph{Baselines}
We choose EmbedKGQA as a baseline.
Our QA method improves EmbedKGQA by replacing its KG embeddings learned by ComplEx \cite{ComplEx} with our transferable embeddings. 
We also choose recent embedding-based QA methods NSM~\cite{NSM_QA_WSDM21} and $\mu$KG \cite{MuKG} as baselines.
To further investigate the effect of KG incompleteness on QA performance,
following EmbedKGQA, we use FB4QA in two settings, i.e., Half-KG and Full-KG.
The former only has half of the triplets in FB4QA,
and the latter is the entire FB4QA.
Our method employs an additional background KG \wdfivem to learn embeddings for FB4QA.
Following EmbedKGQA and NSM, we also use the relation pruning strategy to reduce the space of candidate entities by removing irrelevant relations.

\end{document}